\documentclass{article}

\usepackage{arxiv}
\usepackage[utf8]{inputenc} 
\usepackage[T1]{fontenc}    
\usepackage{url}            
\usepackage{booktabs}       
\usepackage{amsfonts}       
\usepackage{nicefrac}       
\usepackage{microtype}      
\usepackage{hyperref}
\usepackage{lipsum}         
\usepackage{graphicx}
\usepackage{doi}
\usepackage{enumerate}
\usepackage{amsmath}
\usepackage{bbm}
\usepackage{booktabs}
\usepackage[table,xcdraw]{xcolor}
\usepackage{multirow}
\usepackage{svg}
\usepackage{cleveref}       
\usepackage{algorithm}
\usepackage{algpseudocode}

\usepackage{booktabs}
\usepackage{graphicx}
\usepackage[normalem]{ulem}
\useunder{\uline}{\ul}{}

\usepackage[style=ieee, isbn=false]{biblatex}
\AtEveryBibitem{
    \clearfield{urlyear}
    \clearfield{urlmonth}
}
\DefineBibliographyStrings{english}{
  url         = Available ,
}

\title{Generative multi-domain transfer learning for fault detection in data-scarce wind turbines}
\date{}

\newif\ifuniqueAffiliation

\usepackage{authblk}

\author[1, 2]{%
	Stefan Jonas\thanks{\texttt{stefan.jonas@bfh.ch}}}%

\author[1,3]{%
	Angela Meyer}%

\affil[1]{School of Engineering and Computer Science, Bern University of Applied Sciences, Biel, Switzerland}
\affil[2]{Faculty of Informatics, Università della Svizzera italiana, Lugano, Switzerland}
\affil[3]{Department of Geoscience and Remote Sensing, Delft University of Technology, Delft, The Netherlands}

\renewcommand{\headeright}{}
\renewcommand{\undertitle}{}
\renewcommand{\shorttitle}{}

\begin{document}
\maketitle

\begin{abstract}
Normal behavior models have shown promise for reliable fault detection in wind turbines. However, these unsupervised anomaly detection models require sufficient fault-free training data to learn the normal operation behavior of turbines. Under data scarcity, for example in newly deployed wind turbines, these models may result in poor fault detection performance. In this work, we propose a multi-domain generative domain mapping approach based on Star Generative Adversarial Networks (StarGAN) to improve fault detection on data-scarce wind turbines. Our model maps SCADA measurements from a data-scarce turbine to resemble those of several data-rich turbines. By preserving the operational state during translation, faults occurring in a data-scarce domain can be mapped and detected by reliable pre-trained normal behavior models of data-rich domains. Highlighting the benefits of an ensemble fusion strategy, we show that under severe data scarcity our method can produce anomaly scores comparable to models trained on large representative datasets. Our approach can consistently outperform models trained on scarce data when less than 2 weeks of training data are available. With just 2 weeks of accumulated training data, we achieve an anomaly score similarity that is, on average, +16\% higher than conventional fine-tuning, and +10\% higher than single-source domain mapping. As a step towards unsupervised model selection, we propose a proxy metric that detects poor performance at training time, despite an absence of anomalies. Our study presents the potential and challenges of multi-domain mapping for wind turbine fault detection under unrepresentative training data.
\end{abstract}



\section{Introduction}
Early and reliable fault detection in wind turbines (WTs) is crucial for lowering unplanned downtime, reducing maintenance costs, and maintaining efficient operation. Especially deep learning-based models have emerged as a viable approach for continuously monitoring wind turbine behavior \cite{helbingDeepLearningFault2018a}. Fault detection can be framed as a supervised classification problem that distinguishes between labeled normal and faulty turbine data, such as SCADA (Supervisory control and data acquisition) measurements. In practice, this requires an extensive fault database that may be infeasible to obtain. A more practical approach are normal behavior models (NBMs) \cite{wes-8-893-2023}, which instead learn the normal WT behavior only from observations collected during fault-free operation. During operation, NBMs detect deviations from learned feature relationships, which may indicate abnormal behavior and therefore potential faults. NBMs belong to the class of unsupervised anomaly detection, sometimes also termed semi-supervised \cite{ruffUnifyingReviewDeep2021a}. Deep learning-based unsupervised anomaly detection has demonstrated considerable success as NBMs in wind energy (e.g., \cite{meyerMultitargetNormalBehaviour2021a}).
However, these models demand a large dataset of fault-free observations to effectively capture healthy system behavior. Crucially, the training data must be representative, i.e., it must cover the range of operating states that define normal turbine behavior. Under certain circumstances, these demanding data requirements may not be met. For instance, newly commissioned wind farms may have accumulated only a few weeks of historical data, which is often insufficient to capture the full range of operating conditions. Prior literature has shown the detrimental effects of such training data scarcity on anomaly detection performance  \cite{jonasFaultDetectionNew2025, grataloupWindTurbineCondition2025}.

To address this limitation, we employ deep learning-based transfer learning and domain adaptation \cite{panSurveyTransferLearning2010a, wilsonSurveyUnsupervisedDeep2020a}. These methods transfer knowledge from related data-rich domains to improve learning in domains where data is limited. In our context, the source domains are one or more data-rich WTs with representative training data, and the target domain is a WT with only scarce data. Domain adaptation transfers knowledge from the source domain(s) to improve fault detection in the target domain. A key challenge is to account for domain shifts that arise from differences in turbine specifications, environmental conditions, site characteristics, and operational histories. 

While domain adaptation has been extensively studied in supervised settings, it is comparably less explored for unsupervised anomaly detection, in wind energy and more generally \cite{yang2023anomaly}. The few existing applications for this WT fault detection task (\cite{schroderUsingTransferLearning2022a, roelofsTransferLearningApplications2024, zgraggenTransferLearningApproaches2021}) tend to rely on conventional transfer learning strategies such as fine-tuning. Novel domain adaptation techniques suitable for transferring knowledge across vastly different turbines coupled with strong data scarcity remain largely unexplored. A notable exception is domain mapping \cite{wilsonSurveyUnsupervisedDeep2020a}, a generative domain adaptation strategy that has shown early promise. First proposed for this task by Jin et al. \cite{jinConditionMonitoringWind2023} and later improved and extended by Jonas and Meyer \cite{jonasFaultDetectionNew2025}, the objective is to translate SCADA measurements from one WT (domain) so that they resemble those of another. Effectively, a generative model learns to transform data from the scarce target WT to look like data from the data-rich source WT. During operation, target WT data is mapped to resemble the source WT and then evaluated using the source WT's reliable NBM. Performing fault detection in the domain space of a substantially different data-rich turbine has been shown to improve fault detection performance under conditions of severe data scarcity.

Despite this potential, these studies still have notable practical limitations. In this setting, only one data-rich WT is used as the source. In practice, however, representative datasets from several WTs may be available. Such a scenario raises the question of which specific WT should serve as the source domain. As discussed in \cite{jonasFaultDetectionNew2025}, this domain pair selection is an open challenge: The source WT can strongly affect performance, yet cannot be adequately selected. This difficulty is linked to the lack of representative target WT validation data and the inherent absence of anomalies, making it challenging to assess future performance at training time. This limitation generally hinders tuning and evaluation to establish whether a suitable domain mapping model was successfully trained, further contributing to an uncertainty of future reliability.

\begin{figure}[ht!]
    \centering
    \includegraphics[width=0.6\textwidth]{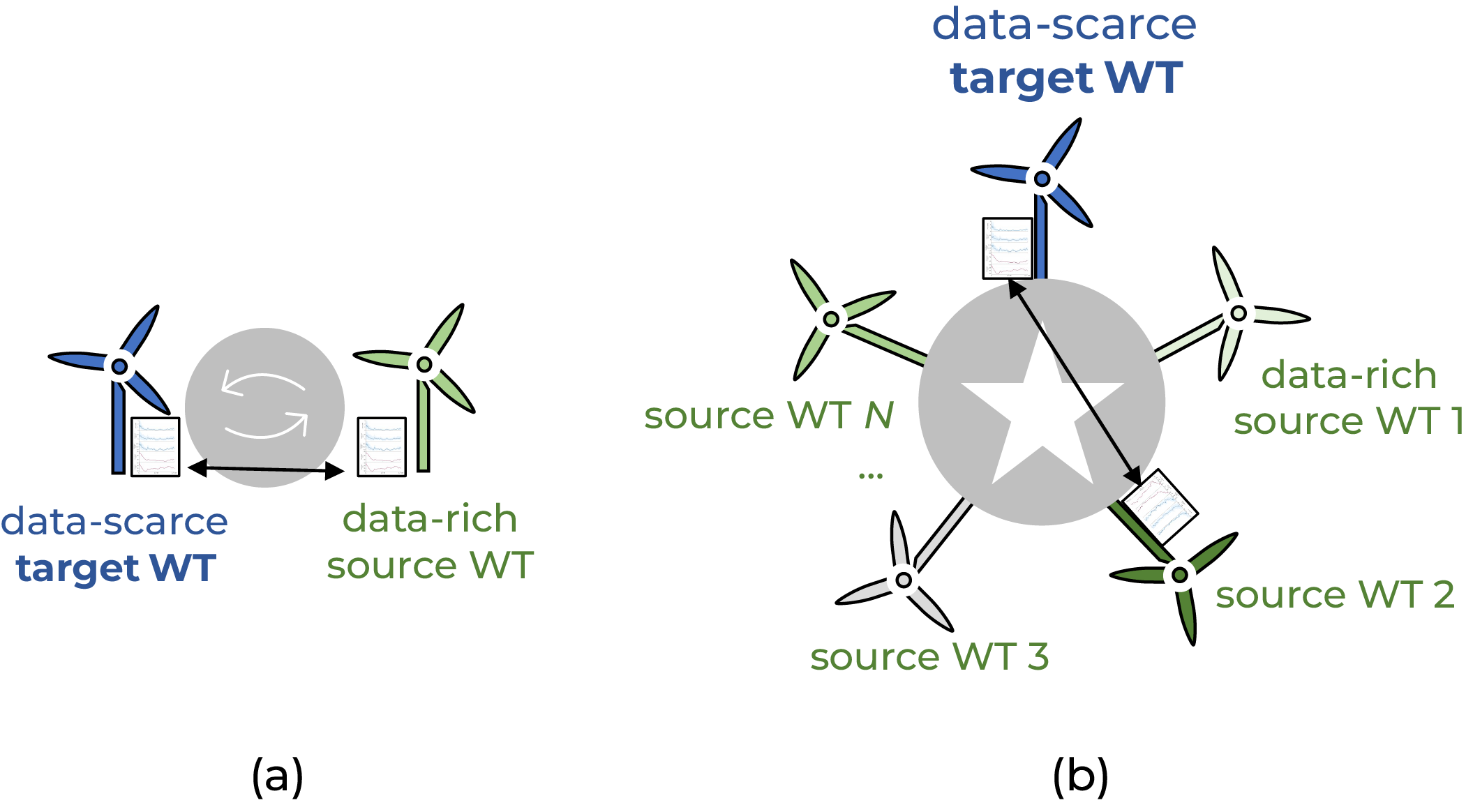}
    \caption{(a): A single-source domain mapping framework, where samples are mapped between one data-scarce target WT and one data-rich source WT. (b): Our multi-source framework mapping samples across several data-rich source WTs and one data-scarce target WT.}
    \label{fig:multi-source}
\end{figure}

To overcome these challenges, we extend the setting to multiple source domains by leveraging information from several source WTs at once. Our proposed approach builds on StarGAN \cite{choiStarganUnifiedGenerative2018}, an unpaired image-to-image translation framework mapping multiple domains to each other with just a single model. Instead of training a separate model for each WT-to-WT pair, we demonstrate that one generative network can map the scarce target WT to all available source WTs simultaneously. We show that we can address open challenges in WT pair selection and model selection, while additionally enabling practical ensemble fusion for more robust performance. Figure \ref{fig:multi-source} contrasts single WT-to-WT mapping with our multi-source domain formulation. StarGAN has previously been adapted to time series data, for instance for EEG signals \cite{KWON2022117574}, speech data \cite{li2021starganv2, emotionrecognitionspeech, meftahEnglishEmotionalVoice2023}), or structural health monitoring \cite{avciATStarGANGPAttentionenhancedMultidomain2026, avciCrossstructureDomainTranslation2026}, but, to our knowledge, not to unsupervised anomaly detection.

\clearpage

The main contributions of this study are as follows.

\begin{enumerate}[i)]
  \item We present a StarGAN-based domain adaptation framework that maps a data-scarce target WT into several data-rich source WTs to improve unsupervised anomaly detection under limited data.
  \item We introduce a novel proxy metric for model selection that identifies, during training, models likely to perform poorly in operation (at inference time), despite the absence of anomalies.
  \item We investigate and evaluate ensemble fusion strategies to combine detection results from all source domains, thereby removing the dependence on selecting a single source turbine.
\end{enumerate}

\section{Related Work}
\label{sec:relwork}
\subsection{Domain adaptation and problem setting}
In single-source domain adaptation, the goal is to learn a machine learning task on a target domain by leveraging a related but different source domain. A domain is denoted as $\mathcal{D}=\{\mathcal{X},P(X)\}$, with its observations $X = \{x_i\}_{i=1}^n$ from the feature space $\mathcal{X}$ sampled from its marginal probability distribution $P(X)$ \cite{panSurveyTransferLearning2010a}. A task $T = \{\mathcal{Y}, P(Y|X)\}$ is defined by the label space $\mathcal{Y}$ and the conditional probability $P(Y|X)$. Knowledge from a source domain $\mathcal{D}_S$ with associated data and label samples $\{(x^s_i, y^s_i)\}_{i=1}^{n_s}$ and task $T_S$ is adapted to learn $T_T$ in a different target domain $\mathcal{D}_T$ with samples $\{(x^t_i, y^t_i)\}_{i=1}^{n_t}$. The key challenge is the distributional difference between the domains, referred to as domain shift (e.g., \cite{Luo_2019_CVPR}). A prominent application is learning to classify unlabeled target domain images ($\{x^t_i\}_{i=1}^{n_t}$ with unknown $\{y_i^t\}_{i=1}^{n_t}$) by utilizing labeled source domain data $\{(x^s_i, y^s_i)\}_{i=1}^{n_s}$.

Our problem setting of domain adaptation for unsupervised anomaly detection \cite{yang2023anomaly} differs from that common setting. First, it is assumed that only normal training data is available. Thus, while all training labels are known in both domains, they all denote a normal sample, i.e., $y^s_i = 0$ , $y^t_j = 0, \forall\, i,j$. Anomalous samples ($y=1$) can only appear at inference time. Moreover, in contrast to the usual availability of abundant target domain samples, our challenge is a general data scarcity in the target domain for both observations and labels, i.e., $n_t \ll n_s$. Lastly, rather than a single source domain, we use \emph{several} source domains $\{\mathcal{D}_{S_i}\}_{i=1}^N$, placing our approach within multi-source domain adaptation (e.g., \cite{SUN201584, NEURIPS2019_db9ad56c}), 

We employ domain mapping (or translation), a generative domain adaptation technique that can transform unpaired samples between source and target domains while preserving their content (e.g., \cite{NIPS2017_59b90e10}). For a general overview of domain mapping, we refer to \cite{wilsonSurveyUnsupervisedDeep2020a} and to \cite{jonasFaultDetectionNew2025} for its applications to WT-related areas. Our goal is to learn a mapping $f(x^t; i): \mathcal{D}_T \rightarrow \mathcal{D}_{S_i}$ from the target to the source domain. The mapped sample is then scored by the source's anomaly scoring function (an NBM in our WT context) $\phi_i: \mathcal{D}_{S_i}\rightarrow \mathbb{R}$, yielding an anomaly score $s = \phi_i(f(x^t; i))$.

\subsection{Transfer learning for WT fault detection}
Very little attention has been devoted to domain adaptation for unsupervised anomaly detection  \cite{yang2023anomaly, michauUnsupervisedTransferLearning2021a}. Only few studies investigated improving WT fault detection with SCADA-based NBMs when faced with data scarcity. Schröder et al. \cite{schroderUsingTransferLearning2022a} present a fine-tuning strategy where an NBM is pretrained on abundant simulation data and then fine-tuned to a real WT with limited samples. Results show that fine-tuning an ANN-based NBM can yield improvements when only 1 month of data is available. However, the approach is constrained by the availability of representative physics-informed simulation data. Three transfer learning strategies were investigated by Zgraggen et al. \cite{zgraggenTransferLearningApproaches2021} to adapt a regression-based source WT NBM to a target WT from a different farm: A linear regression-based correction of the source NBM output, an extension to it with a CNN modeling the remaining error component, and fine-tuning. Using a comparably large target dataset comprising 3 months, the CNN-extended correction strategy is shown to result in similar performance as a standalone model trained on abundant target data. For fine-tuning, its proneness to forget source knowledge and the importance to investigate the potential effect of selecting an appropriate source WT in the future are highlighted. Roelofs et al. \cite{roelofsTransferLearningApplications2024} present a transfer learning approach to tune an autoencoder-based source NBM to turbines from the same wind farm. Fine-tuning using 1 to 3 months of target WT data was shown to perform as well or slightly better than NBMs trained on abundant data. Notably, a multi-asset strategy, involving pretraining one NBM on multiple source WTs, did not yield improvements. Transferring within the same farm is likely to be less challenging due to a smaller domain shift between WTs and instead mainly aims to achieve more efficient training of fault detection models across large fleets. 
\clearpage
\subsection{Advanced domain adaptation strategies for WT fault detection}
A novel approach was presented by Jin et al. \cite{jinConditionMonitoringWind2023} to learn a feature mapping from a data-scarce target WT to a substantially different source WT. A generative adversarial network is trained to transform target WT data into data matching the distribution of the source WT. At test time, the model maps target WT observations to the source WT, which are evaluated for faults using the reliable source NBM. Results from two case studies show improved fault detection compared to a target NBM trained on severely limited training data. 

Jonas and Meyer \cite{jonasFaultDetectionNew2025} extended this to a content-preserving domain mapping approach. A CycleGAN\cite{zhuUnpairedImagetoimageTranslation2017}-based domain mapping with content-preservation losses is proposed to preserve the content of SCADA samples across domains. Using the cycle-consistency and introducing physics-informed losses, the operational state such as idle or maximum power generation are preserved from one WT to another. Results across combinations of target and source WTs show on average better anomaly scores compared to NBMs trained on scarce data and fine-tuning when faced with a severe data scarcity corresponding to only one week to one month of training data. However, a large variability across domain pairs is noted, highlighting the importance of selecting an appropriate source WT and assessing future model performance at training time. 

\section{Dataset}

\subsection{Wind turbines and SCADA data}
Our study utilizes SCADA measurements from seven operational WTs collected from different onshore wind farms across separate locations. The dataset comprises 10-minute averaged SCADA measurements collected during a multi-year period. All turbines are from the same manufacturer and share identical SCADA system variables, but can vary in model specifications and rated power. Each turbine's dataset is split into a training set, comprising the first 70\% of measurements, and a test set, comprising the remaining 30\% of data. The last 30\% of the training set is held out as a validation set. Unsupervised anomaly detection assumes that only fault-free samples are present at training time. To uphold this assumption, we combine a rated power filter with a procedure based on the Mahalanobis distance \cite{mckinnonComparisonNovelSCADA2022} to exclude curtailments, outliers, and potentially anomalous measurements from training and validation sets. Our dataset does not include fault labels but time periods flagged as incidents for which no further details are provided. We exclude measurements recorded during periods flagged as containing an unspecified incident, although this flag does not necessarily indicate faulty operation. No filtering is applied to the test set, which remains uncleaned and can therefore contain (unlabeled) abnormal samples, reflecting real operating conditions.

The datasets containing 10-minute SCADA averages are then converted into 12-hour time series using a sliding window technique. Each sample comprises channels characterizing the wind speed, rotor speed, power, mean stator temperature, and rotor temperature. The wind-, rotor speed, and power output are represented by three features each representing their minimum, average, and maximum values during the 10-minute period, resulting in a total of 11 channels for each sample.

Our work follows the same dataset splitting and processing procedure as presented in \cite{jonasFaultDetectionNew2025}. Details about the turbines are provided in Table \ref{tab:datatable}.

\subsection{Defining source and target domains}
In our study, we aim to leverage data from several source domains, represented by data-rich wind turbines, to improve fault detection on a target domain, represented by a data-scarce wind turbine. We define six different setups, each representing a different assignment of a target WT and several source WTs. In setup $x$, turbine WT0$x$ is designated as the target WT, while a selection of the remaining turbines act as data-rich source WTs. We define six setups, as WT07 was excluded from experiments, following \cite{jonasFaultDetectionNew2025}, because it was used for model selection in that study. The corresponding source domains within each setup were chosen such that the target WT operationally differs from every source WT within a setup. 

\begin{figure}[ht!]
    \centering
    \includegraphics[width=.51\textwidth]{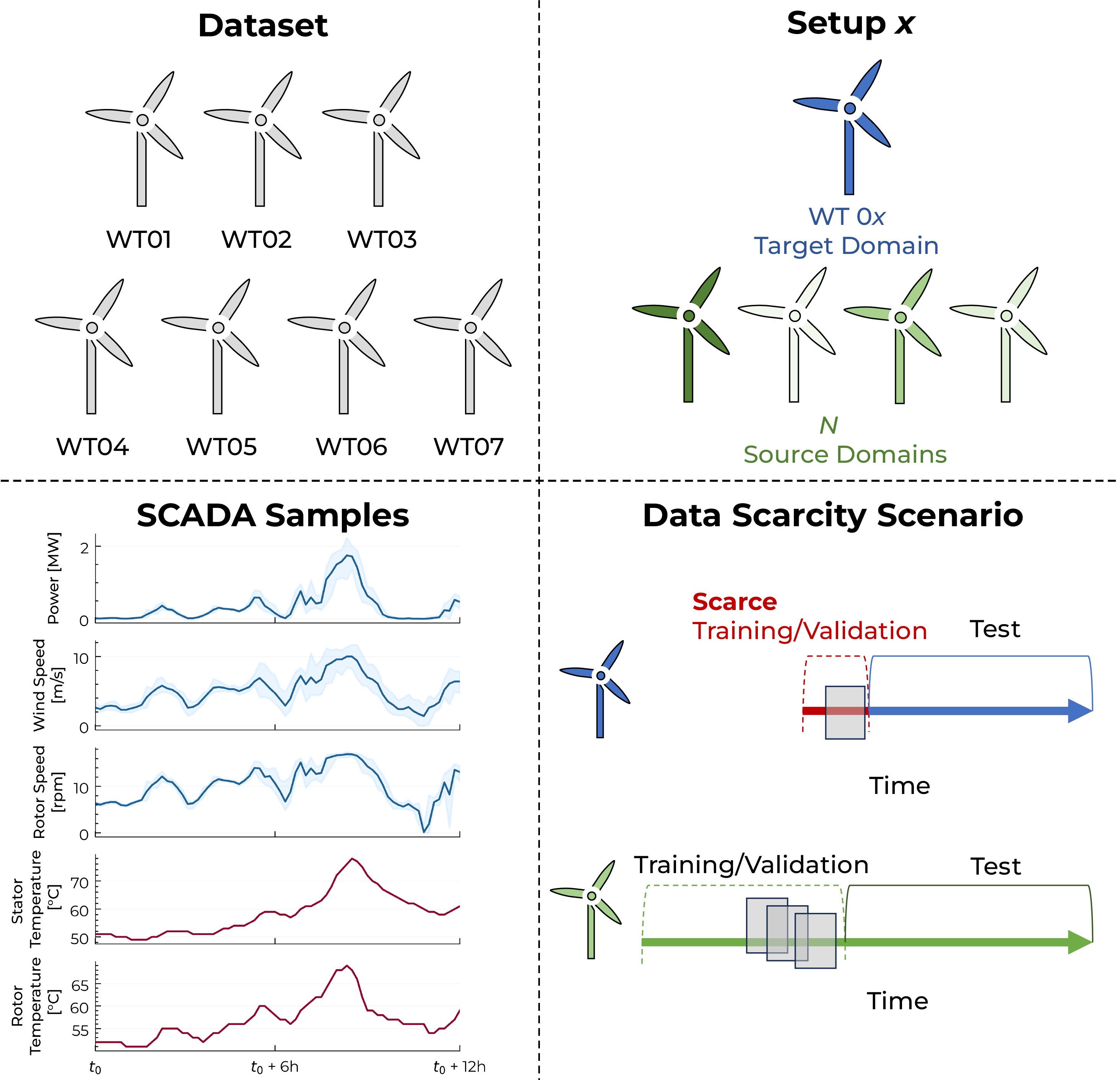}
    \caption{Visual summary of our dataset, comprising 7 different WTs, which are organized into setups of a selected target WT and several source domain WTs. The target WT has limited training data available, while the data-rich source WTs provide representative training data. Our study works with SCADA samples spanning 12 hours.}
    \label{fig:dataset}
\end{figure}

Once a setup specified a target WT, an artificial data scarcity scenario is applied to its training and validation sets. Specifically, only 1 week up to 2 months of SCADA samples (1008 to 8064 samples) observed immediately prior to the unaffected test set are retained from the target domain's training and validation set. Lastly, within each setup we applied channel-wise min-max normalization to scale all channel values to the range $[-1, 1]$ by using the training set statistics of the first source WT. All data processing steps are further described and available in our provided implementation (see Appendix B). Our dataset, samples, and multi-domain transfer learning setup are visualized in Figure \ref{fig:dataset}.

\section{Methodology}
\begin{figure}[ht!]
    \centering
    \includegraphics[width=0.85\textwidth]{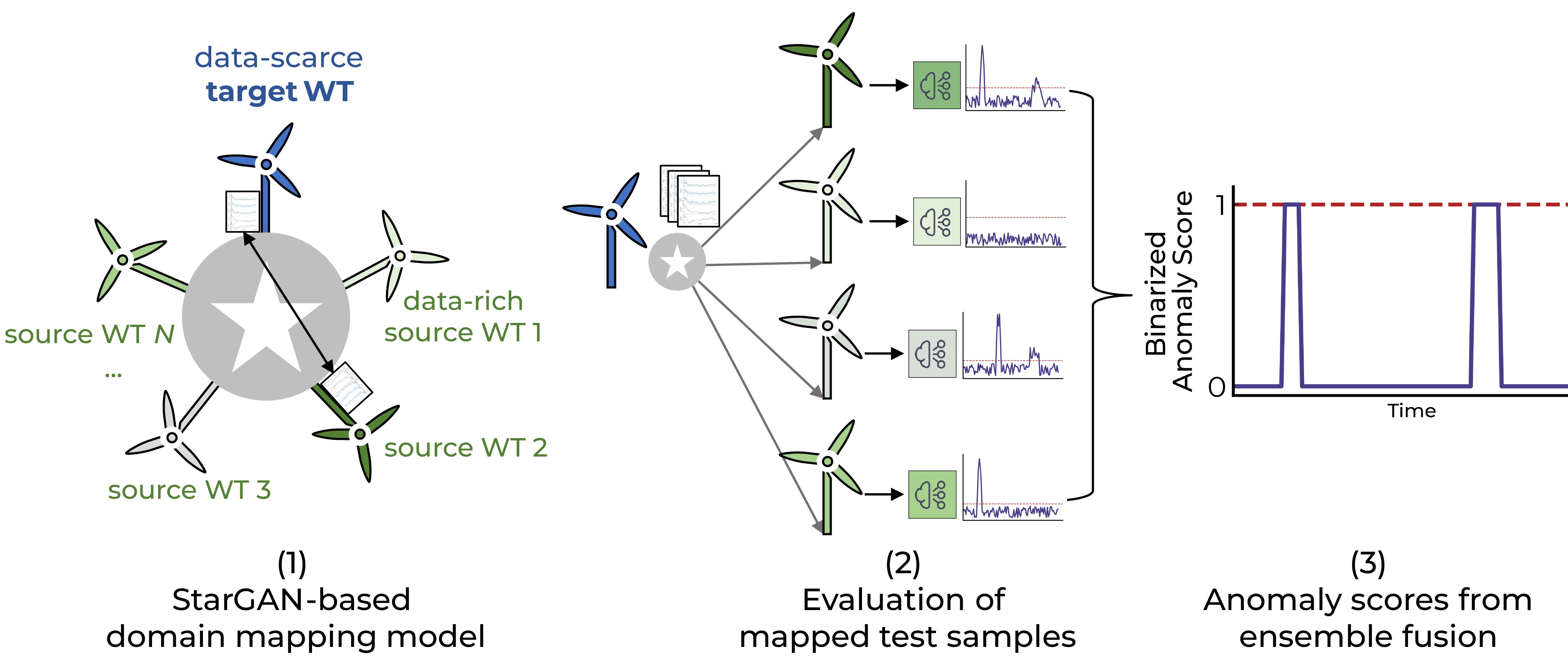}
    \caption{A visual summary of our proposed approach. (1): A StarGAN model is trained to map SCADA samples across several source WTs and a target WT. (2): At inference time, target WT test data is mapped to all source WTs and evaluated with their corresponding pretrained NBMs. (3): Anomaly scores are binarized and combined using an ensemble fusion strategy to obtain the final scores.}
    \label{fig:methodology}
\end{figure}

Training normal behavior models on non-representative training data may lead to inaccurate fault detection. We propose to leverage representative training data from several different WTs to mitigate this performance deterioration on a data-scarce WT. We employ domain mapping, a generative transfer learning strategy, to accomplish this goal. Our presented generative network learns to map (translate) fault-free SCADA samples across WTs by transforming the sensor measurements of one WT to those of another. The mapping retains characteristics of the operational state across turbines. During operation (at inference time), SCADA data originating from the data-scarce WT (target domain) is mapped to all data-rich WTs (source domains), where translated anomalous states can be detected using their reliable normal behavior models. We further take advantage of our multi-source formulation by applying an ensemble fusion strategy to combine several predictions for the data-scarce WT. The general framework is shown in Figure \ref{fig:methodology}. 

\subsection{Domain mapping network}
\subsubsection{Multi-domain mapping with StarGAN}
We build our domain mapping network upon StarGAN \cite{choiStarganUnifiedGenerative2018}, a framework designed for unpaired image-to-image translation across $N$ domains. StarGAN consists of only a single shared generator $G$ and a single discriminator $D$ to learn all domain mapping directions simultaneously. In its originally proposed formulation, the StarGAN generator takes an input image $x$ from an origin domain coupled with a provided label $c$, defining a destination domain, to generate an output image $y$ that resembles an image of the destination domain, i.e., $G(x, c) = y$. 

StarGAN is trained by optimizing an objective consisting of multiple losses. First, to generate realistic samples, the generator and discriminator follow a GAN framework defined by an adversarial loss, i.e. $\mathcal{L}_{adv} = \mathbb{E}_x [\log D_{src}(x)] + \mathbb{E}_{x,c}[\log(1-D_{src}(G(x, c))]$, where $D_{src}$ represents a probability distribution over sources given by $D$. 

As the goal is to generate samples belonging to a specific domain, a classification loss $\mathcal{L}_{cls}$ is introduced through a classifier head added to $D$. The discriminator is trained to classify real samples $x$ according to their corresponding origin domain label $c'$, i.e., $\mathcal{L}^r_{cls} = \mathbb{E}_{x, c'}  [-\log D_{cls}(c'|x)]$. The classifier output is employed to encourage the generator to generate samples matching a particular destination domain $c$, based on the classification loss of fake samples: $\mathcal{L}^f_{cls} = \mathbb{E}_{x, c}  [-\log D_{cls}(c|G(x, c)]$. 

StarGAN is based on unpaired image-to-image translation that aims to preserve input content during a translation by only changing domain-specific parts of an image. A cycle-consistency loss addresses this by requiring translations to be invertible, enforcing that no domain-invariant features are removed, added, or altered. Formally, $\mathcal{L}_{cyc}= \mathbb{E}_{x, c, c'} [\| x - G(G(x, c), c')\|_1]$ is added to the generator objective, pushing it towards small reconstruction errors when translating a sample from an origin domain labeled $c'$ to a destination domain labeled $c$ and back. 

The full objective is defined as:
\begin{equation}
    \mathcal{L}_D = -\mathcal{L}_{adv} + \lambda_{cls}\mathcal{L}^r_{cls},
\end{equation}
\begin{equation}
    \mathcal{L}_G = \mathcal{L}_{adv} + \lambda_{cls}\mathcal{L}^f_{cls} + \lambda_{cyc}\mathcal{L}_{cyc},
\end{equation}

where $\lambda_{cls}$ and $\lambda_{cyc}$ are tunable hyperparameters defining the trade-off between sample realism, domain classification, and content preservation.

\subsubsection{StarGAN for WT mapping}
We modify StarGAN to adapt it from multi-domain image-to-image translation to our objective of mapping SCADA time series samples between $N$ source WTs and one data-scarce target WT. All $N+1$ domains are represented by WTs, with each data-rich source WT possessing a representative training set, and the data-scarce target WT being affected by a training data scarcity scenario.

We add two physics-informed content-preservation losses, namely a zero loss $\mathcal{L}_0$ and a rated power loss $\mathcal{L}_{\mathcal{R}}$, to the objective. These losses have been shown to contribute towards preserving the operational state during mapping \cite{jonasFaultDetectionNew2025}. The zero loss ensures that idle states - during which the wind turbine generates no power – remain idle during translation. It punishes deviations from a zero state in the mapped output at positions where selected input sample channels are zero-valued, e.g., when the power output was at zero. The rated power loss, on the other hand, encourages the generator to map power values matching the rated power capacity of the origin turbine to the corresponding destination turbine's capacity. 

Formally, for an input sample $x \in \mathbb{R}^{C \times T}$ from an origin domain labeled $o$, with $C$ channels and a length of $T$, we define masks selecting positions relevant for the zero and rated power loss. For the zero loss, we punish deviations from idle states for all power and rotor speed channels ($C_0$) in the masked output sample $G(x, d)$ mapped to a destination domain labeled $d$:

\begin{equation}
    M^0_{c, t} = \mathbf{1}[c \in C_0] \cdot \mathbf{1}[x_{c,t} = 0] 
\end{equation}
\begin{equation}
\mathcal{L}_0 = \text{MAE}(M^0_{c,t}\ G(x, d)_{c,t} - 0)
\end{equation}

For the rated power loss, we compute deviations from the destination turbine's rated power ($R_d$) in relevant positions where the average power channel ($C_R$) matched the origin turbine's rated power ($R_o$):

\begin{equation}
    M^R_{c, t} = \mathbf{1}[c \in C_R] \cdot \mathbf{1}[x_{c,t} = R_o]
\end{equation}
\begin{equation}
    \mathcal{L}_R = \text{MAE}(M^R_{c,t}\ G(x, d)_{c,t} - R_d)
\end{equation} 

Additionally, we adopt the GAN-QP framework \cite{suGANQPNovelGAN2018} as in \cite{jinConditionMonitoringWind2023} and \cite{jonasFaultDetectionNew2025}. Following the latter, we also add an anomaly augmentation step to the training loop, in which randomly selected channels of input samples are artificially corrupted (set to zero). The generator is asked to map corrupted batches back and forth across domains to hinder it from learning to repair samples, i.e., to overfit on generating normal samples only, which would be detrimental to preserve anomalous states at operational time. 

\begin{figure}[ht!]
    \centering
    \includegraphics[width=0.85\textwidth]{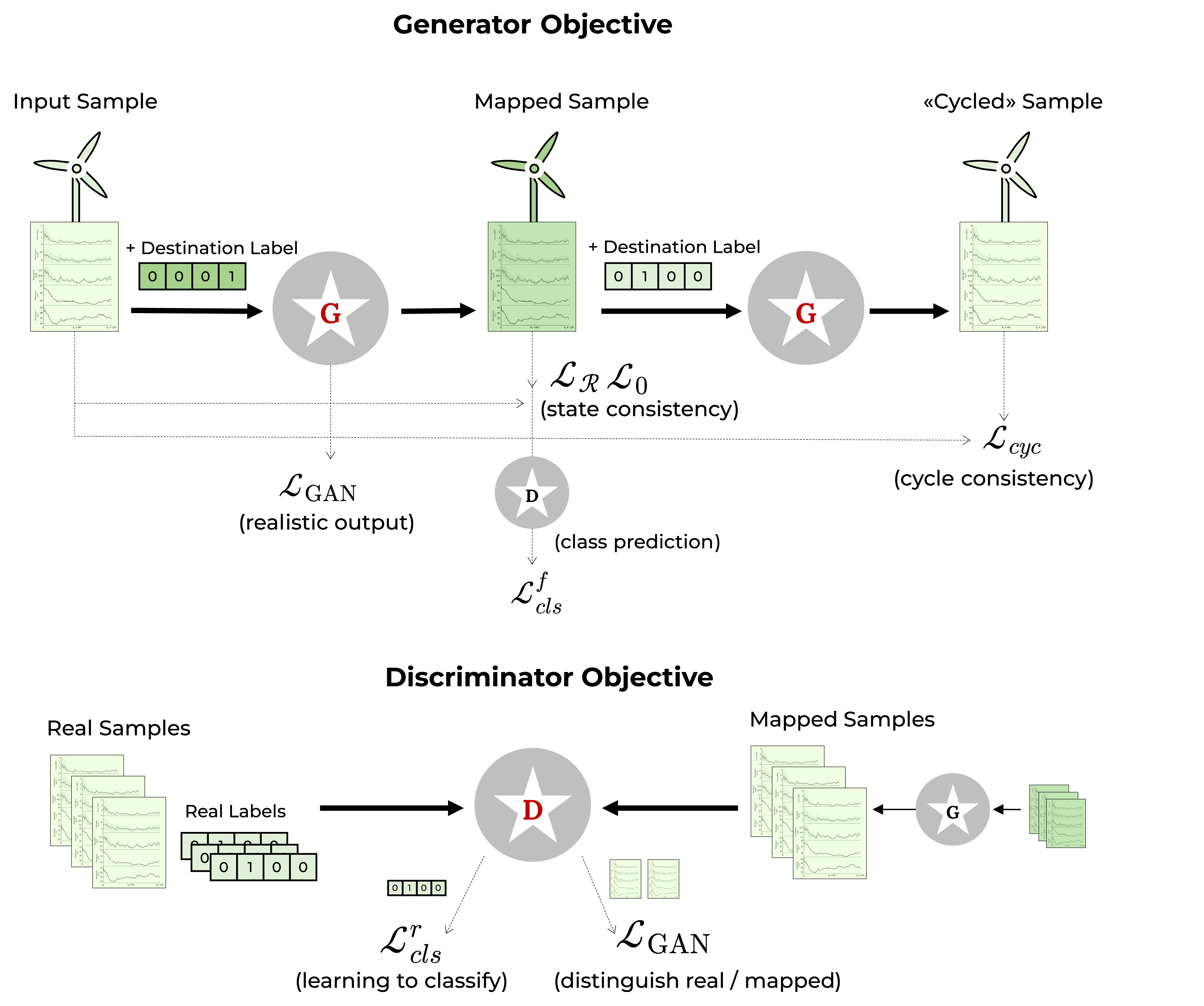}
    \caption{The generator objective is shown at the top. The generator is tasked to translate a sample to the specified destination domain. The adversarial loss forces the generator to output a realistic SCADA sample. It learns to match the destination domain through the classification loss obtained from the discriminator, while the two physics-informed state-consistency losses aim to preserve the operation by matching idle and maximum power states. The cycle-consistency loss directs the generator to map the translated sample back to its original input, enforcing further content preservation. \newline The discriminator objective is shown at the bottom. The discriminator learns to classify real SCADA samples to their corresponding domain and to distinguish between real and mapped samples.}
    \label{fig:objectives}
\end{figure}

Our full objective is shown in Figure \ref{fig:objectives} and defined as :
\begin{equation}
    \mathcal{L}_D = \mathcal{L}_{\text{GAN-QP}_D} + \lambda_{cls_D}\mathcal{L}^r_{cls},
\end{equation}

\begin{equation}
    \mathcal{L}_G = \mathcal{L}_{\text{GAN-GP}_G} + \lambda_{cls_G}\mathcal{L}^f_{cls} + \lambda_{cyc}\mathcal{L}_{cyc} + \lambda_{0}\mathcal{L}_{0}+ \lambda_{\mathcal{R}}\mathcal{L}_{\mathcal{R}},
\end{equation}

where $\{\lambda_{cls_D}, \lambda_{cls_G}, \lambda_{cyc},  \lambda_{0}, \lambda_{\mathcal{R}} \}$ are the loss weights. The selection of these weights is discussed in \ref{sec:model_tuning}.

We use a 1D temporal convolutional network (TCN \cite{baiEmpiricalEvaluationGeneric2018}) with residual blocks for the generator $G$ architecture to process SCADA samples. Its input and output are 12-hour SCADA samples with 11 channels (i.e., of shape $(\text{batch size}, 72, 11)$). The discriminator is a 1D convolutional network taking as input a SCADA sample. Its outputs are a score for the GAN-GP loss and a domain class prediction. Further details on model architectures, training procedure, and implementation are described in Appendix B.

\clearpage
\subsection{Fault detection with domain mapping}
For our downstream fault detection task, we ultimately map test samples from the target WT to all other source WTs for subsequent evaluation. At inference time, the trained generator maps $k$ test set samples from the scarce WT to $N$ source domains, where they are scored using the reliable NBMs pretrained on the data-rich source domains.

\paragraph{Evaluation with NBMs.}
We pretrain separate normal behavior models for all WTs using representative training sets. We employ autoencoder-based NBMs that aim to reconstruct SCADA samples provided as input. The reconstruction error between input and output represents the anomaly score. The NBMs have been trained exclusively on normal data, such that we expect anomalous samples to be reconstructed worse compared to fault-free samples. For each NBM, a threshold $T^*$ differentiating between normal and anomalous is set based on the errors on fault-free validation data. We describe the NBM architecture and training process further in Appendix B.

\paragraph{Ground truth fault detection performance.} 
As our dataset lacks labeled faults, we instead define a ground truth from an NBM trained on representative target WT training data, i.e., without data scarcity. 
We evaluate $k$ test set samples to obtain anomaly scores $s^* = (s^*_1,..., s^*_k)$. We are particularly interested in whether a sample is considered anomalous, that is, if its score exceeds the NBM threshold $T^*$, or if it is normal. We thus define binarized ground truth scores $b^* = (s^*_1 \geq T^*,...,s^*_k \geq T^*)$, with positive $(P, 1)$ and negative $(N, 0)$ values indicating anomalous and normal samples, respectively. We assume that the NBM trained on representative training data sufficiently captures true anomalies to act as ground truth.

\paragraph{Anomaly scores from mapped samples.} 
The goal of our domain adaptation approach is to improve fault detection under data scarcity. To this end, we assess our model's performance by comparing whether it produces anomaly scores that are similar to those of the ground truth, obtained without data scarcity. Mapping $k$ test set samples from the target WT and scoring them in the source domains using their NBMs, $\text{NBM}_d$, results in $N$ anomaly scores $s^d = (s^d_1,...,s^d_k)$ for each source domain $d \in \{1,...,N\}$. All scores are subsequently binarized based on the corresponding threshold, $T$, of $\text{NBM}_d$, i.e., $b^d = (s^d_1 \geq T^d,...,s^d_k \geq T^d)$.

\paragraph{Ensemble fusion strategies.}
\label{sec:ensemble}
It remains a challenge to select the best mapping direction for the downstream anomaly detection task, as its test performance on anomalies cannot be assessed at training time. We address this through an ensemble fusion strategy that combines the binarized scores ($b^d$) of \emph{all} $N$ source domains into \emph{one} fused score $b^+$. We consider three strategies:

\begin{enumerate}[i)]
  \item Consensus voting: An anomaly is predicted only if all $N$ source NBMs agree.
  \item Majority voting (Anomaly-biased): A sample is considered anomalous if \emph{at least} the majority 
  $(\frac{N}{2})$ of NBMs agree.
  \item Minority voting: A sample is considered anomalous if any of the $N$ NBMs predicts an anomaly.
\end{enumerate}

\paragraph{Ground truth similarity.} 
For the final fault detection assessment, we calculate an F1 score quantifying a similarity between the binarized ground truth scores $b^*$ and the binarized and fused model candidate scores $b^+$ from our proposed approach:

\vspace{0.05cm}
\begin{equation}
    \text{F1-Score} = \frac{2\text{TP}}{2\text{TP}+\text{FP}+\text{FN}}
\end{equation}

\begin{figure}[ht!]
    \centering
    \includegraphics[width=.825\textwidth]{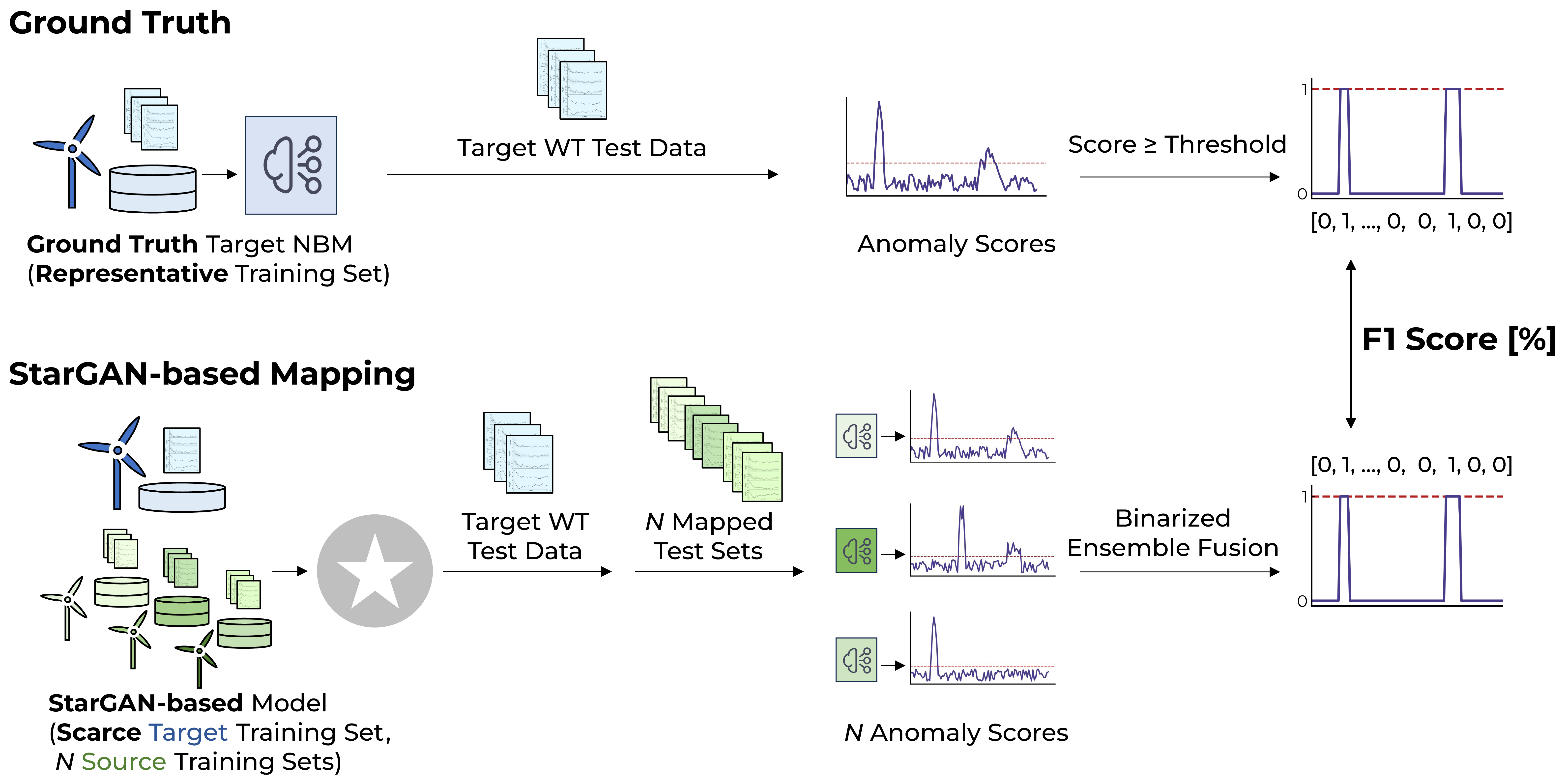}
    \caption{The ground truth anomaly scores are shown at the top. An NBM trained on representative target WT training data evaluates target test samples and yields anomaly scores that are subsequently binarized. We assess the fault detection of our model by mapping target test samples to all source domains for evaluation with their NBMs, before they are binarized and fused, as shown at the bottom. The F1 score is calculated between the binarized ground truth and the fused binarized model scores.}
    \label{fig:evaluation}
\end{figure}

Our fault detection procedure is illustrated in Figure \ref{fig:evaluation}. 

\subsubsection{Unsupervised model selection}
\label{sec:model_tuning}
\paragraph{Model selection.}  Hyperparameters, such as the loss weights, may significantly affect model performance. In conventional supervised machine learning, they are typically set based on validation data representative of test data. However, in our setting, training and validation data comprise only normal samples, significantly impeding model selection and hyperparameter optimization. Therefore, it is challenging to assess and compare operational performance without anomalies at training time. Unsupervised model selection (or evaluation and tuning) is a general challenge across related domain adaptation areas (e.g., \cite{yangCanWeEvaluate2024}). In single-source domain adaptation for WT fault detection, this is particularly challenging, as only one representative validation set is available. The selection of hyperparameters remained unaddressed in Jin et al. \cite{jinConditionMonitoringWind2023}, while in Jonas and Meyer \cite{jonasFaultDetectionNew2025} tuning was performed on a separate test set from an excluded WT, thereby departing from the assumptions of unsupervised anomaly detection and requiring additional data in practice.

Our multi-source framework involves representative validation sets from $N$ data-rich source WTs. We hypothesize that assessing how well these samples are mapped can support unsupervised model selection with normal data. We introduce a proxy metric to capture mapping quality evaluated on fault-free data that is yet reflective of operational performance. Our metric is based on the alignment of real and mapped anomaly scores of normal validation data. Our reasoning is that a good translation model will generate samples that result in similar anomaly scores to origin samples: Subpar models could map all SCADA inputs to a small set of output samples, yielding a concentration of specific anomaly scores. They might also overfit and generate only known samples memorized during training, producing very low anomaly scores. Poor mappings may in contrast cause significantly higher scores. Therefore, we propose to capture the discrepancy between anomaly scores by comparing the relative mass difference in the right tails of the real and mapped score distribution.

Let $\mathcal{D}_{S_i}$ and $\mathcal{D}_{S_j}$ be two data-rich source domains. Given a WT translation $\mathcal{D}_{S_i} \rightarrow \mathcal{D}_{S_j}$, we compare anomaly scores of (normal) validation data from $WT_j$ evaluated with $\text{NBM}_j$ to those scored by mapped validation data from $WT_i$ to $WT_j$. We compare the relative difference in right tail mass, defined as the sum of scores above the $95^{th}$ percentile, $\tau_{95}$, set by the real scores. The right tail mass of the real anomaly scores,
\vspace{0.12cm}
\begin{equation}
    T_{\mathcal{D}_{S_j}} = \sum_{x \in \mathcal{D}_{S_j}} \mathbf{1}[\text{NBM}_j(x) \geq \tau_{95}]
\end{equation}

is compared to the sum of scores above the same threshold with mapped validation samples:
\vspace{0.12cm}
\begin{equation}
    T_{\mathcal{D}_{S_i} \rightarrow \mathcal{D}_{S_j}} = \sum_{x \in \mathcal{D}_{S_i}} \mathbf{1}[\text{NBM}_j(G(x, c_j) \geq \tau_{95}]
\end{equation}

Our proxy metric $\Delta_T$ represents their relative tail difference:
\vspace{0.12cm}
\begin{equation}
    \Delta_T = \frac{|T_{\mathcal{D}_{S_j}} - T_{\mathcal{D}_{S_i} \rightarrow \mathcal{D}_{S_j}}|}{T_{\mathcal{D}_{S_j}}}
\end{equation}

\begin{figure}[ht!]
    \centering
    \includegraphics[width=0.55\textwidth]{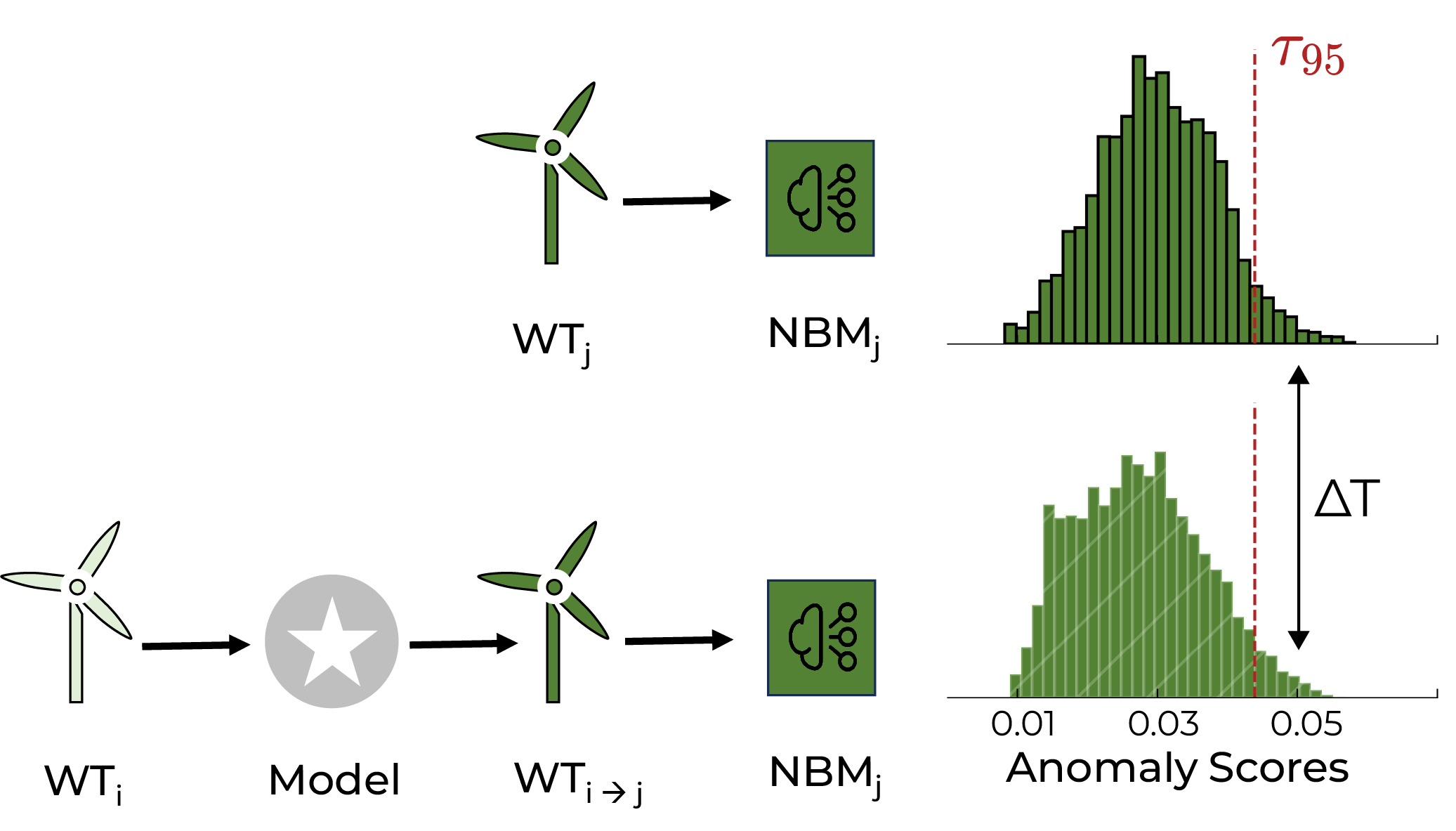}
    \caption{The proxy metric $\Delta_T$ is calculated by comparing the tail of anomaly scores of real validation data (top) and mapped anomaly scores (bottom).}
    \label{fig:deltaT}
\end{figure}

The calculation is visualized in Figure \ref{fig:deltaT}. Model selection is performed by selecting the lowest average tail mass difference across all source-to-source directions with various model configurations. 

\paragraph{Early stopping.} Our proxy metric assesses the mapping quality across source domains. However, an early stopping criterion based on $\Delta_T$ could lead to stopping points focused on improving source-to-source mappings while the model may be overfitting on scarce target WT samples. We instead adopt the early stopping criteria based on the anomaly scores of normal target WT validation data \cite{jonasFaultDetectionNew2025}. At regular checkpoints, the target validation samples are mapped to all source domains and scored with source NBMs. We track the average anomaly score across all directions and stop training once it increases. The anomaly scores decrease in early training as the model learns to generate realistic samples, while overfitting on scarce training samples can lead to validation data being mapped less realistically, causing a rise in average scores. However, this criterion cannot be employed for model selection. As a trivial example, a poor translation model mapping all samples to one same normal SCADA output would result in misleadingly very low criterion score. 
\newline 


\section{Results and Discussion}
We assess whether our proposed multi-source domain mapping can improve fault detection on a data-scarce target domain WT. For each setup, assigning one data-scarce target WT and $N$ data-rich source WTs, we map SCADA observations from the target WT's test set to each source domain, where it is subsequently evaluated for anomalies with the source WT's pretrained NBM. We report the resulting fault detection performance across setups and data scarcity scenarios, compare it to single-source domain adaptation techniques, and investigate the effectiveness of our proxy metric for model selection.
\subsection{Baselines}
We compare the improvement in fault detection performance of our multi-source framework to 3 different baselines: 
\vspace{-5pt}
\paragraph{Data-scarce NBMs.} The fault detection performance is obtained by training an NBM on the scarcity-affected training set of a target WT. The test data is scored and binarized for comparison. This baseline represents the model performance when no transfer learning strategy is employed. 
\vspace{-5pt}
\paragraph{Fine-tuning.} We also compare to fine-tuning, a conventional transfer learning approach. A pretrained NBM from a source WT is slightly adjusted to the target domain by updating its weights through additional learning updates with only scarce target data. We extend fine-tuning to our multi-source setting by fine-tuning several source NBMs to obtain anomaly scores for each source-to-target direction, i.e., a total of $N$ different results for each setup, which are combined into a final score according to an ensemble fusion strategy. 
\vspace{-5pt}
\paragraph{Single WT-to-WT mapping.} Lastly, we consider CycleGAN-based single WT-to-WT domain mapping. The single-source framework is adjusted to several source domains by training separate models to map data between the target WT and each source WT. $N$ domain mapping models translating the target domain data to each source domain were trained. The mapped data is scored using the corresponding source domain's NBM and fused across all source domains.

\begin{figure}[ht!]
    \centering
    \includegraphics[width=.825\textwidth]{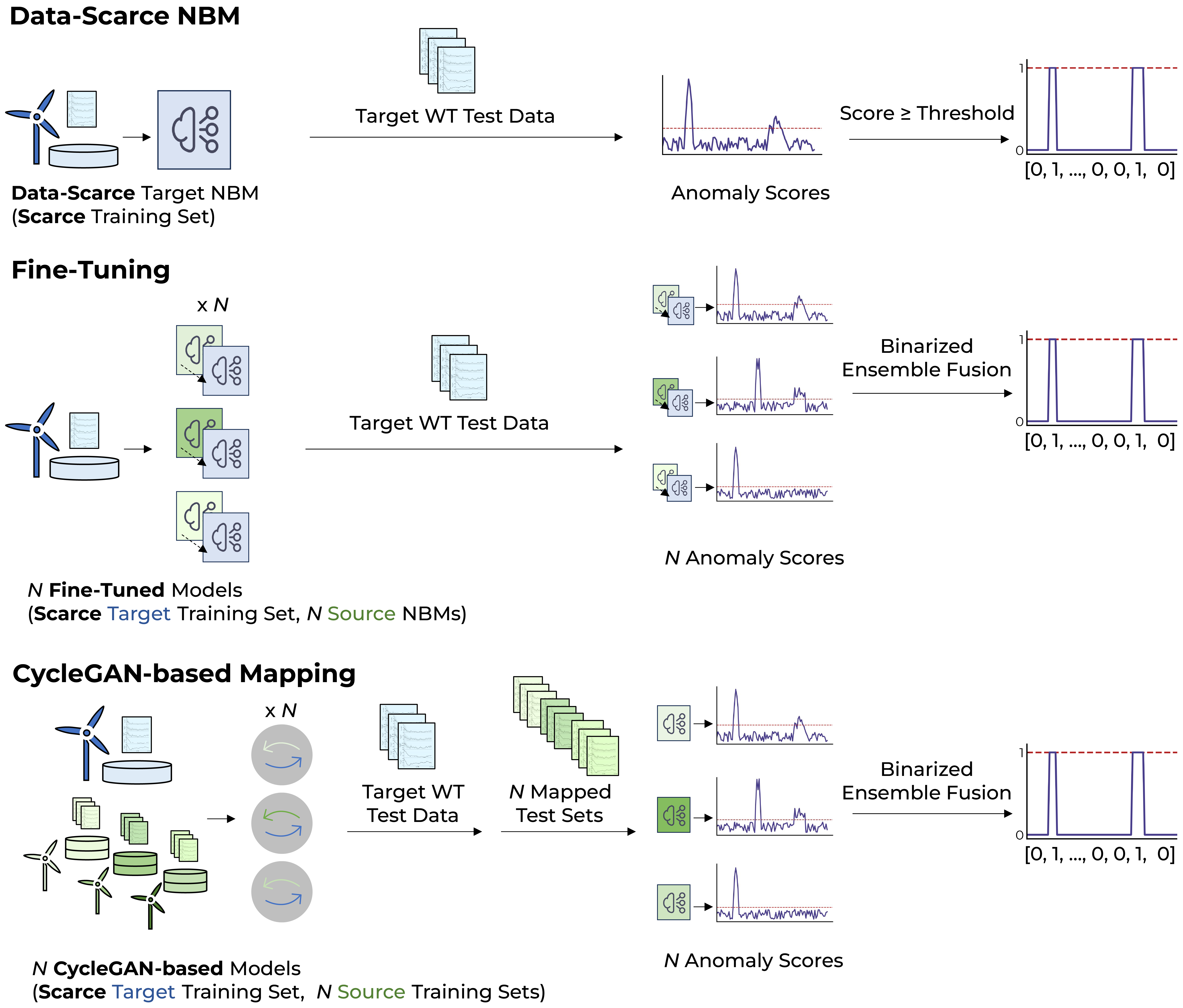}
    \caption{We assess the baseline performances through the binarized anomaly scores of an NBM trained on scarce training data (top), the fused binarized anomaly scores of $N$ fine-tuned source domain NBMs (middle), and the binarized scores from $N$ single-source domain mapping models (bottom).}
    \label{fig:baselines}
\end{figure}

All baselines follow the implementations and models from \cite{jonasFaultDetectionNew2025} and use its pretrained models. The baselines are illustrated in Figure \ref{fig:baselines}.

\subsection{Ensemble fusion}
The fine-tuning and CycleGAN-based domain mapping baselines are originally based on single-source domain adaptation. In our study, we instead assume several source domains - a natural scenario in wind farms or, more generally, fleets of similar devices. In this multi-source setting, these methods result in $N$ separate models and anomaly scores for the target WT. While our StarGAN-based approach is a single shared generator, it nonetheless maps target test data separately to $N$ source WTs. Each of these $N$ mapping directions can result in significantly different fault detection performance on their own. Given that the optimal selection from $N$ source-target pairs remains challenging, we evaluated three different ensemble fusion strategies to obtain a combined output. As shown in Table \ref{tab:ensemble}, the anomaly-biased majority voting outperforms both consensus and minority voting for combining N anomaly scores. \begin{table}[htb!]
\centering
\caption{Changes in F1 score compared to the worst, average, and best score out of $N$ binarized anomaly scores when applying an ensemble fusion strategy. Average improvements are shown across all setups and data scarcity scenarios.}
\label{tab:ensemble}
\resizebox{\textwidth}{!}{%
\begin{tabular}{@{}lccclccclccc@{}}
\toprule
\multirow{2}{*}{\begin{tabular}[c]{@{}c@{}} \end{tabular}} &
  \multicolumn{3}{c}{Fine-tuning} &
   &
  \multicolumn{3}{c}{CycleGAN-based mapping} &
   &
  \multicolumn{3}{c}{StarGAN-based mapping (Ours)} \\ \cmidrule(lr){2-4} \cmidrule(lr){6-8} \cmidrule(l){10-12} 
 &
  Consensus &
  Majority V.&
  Minority V.&
   &
  Consensus &
  Majority V.&
  Minority V.&
   &
  Consensus &
  Majority V.&
  Minority V.\\ \midrule
\multirow{2}{*}{$\Delta$F1 vs. worst}   & 8.7          & \textbf{15.2}  & 12.7         &  & 13.5         & \textbf{15.4} & 3.8          &  & 4.6          & \textbf{19.6} & 14.7         \\
                         & \small{{[}$\pm$14.7{]}} & \small{{[}$\pm$13.2{]}}   & \small{{[}$\pm$17.6{]}} &  & \small{{[}$\pm$15.3{]}} & \small{{[}$\pm$12.1{]}}  & \small{{[}$\pm$9.3{]}}  &  & \small{{[}$\pm$7.0{]}} & \small{{[}$\pm$17.9{]}}  & \small{{[}$\pm$20.1{]}} \\
\multirow{2}{*}{$\Delta$F1 vs. average} & -4.1         & \textbf{2.5}   & -0.1         &  & 0.7          & \textbf{2.6}  & -9.0         &  & -9.1         & \textbf{5.8}  & 0.9          \\
                         & \small{{[}$\pm$14.9{]}} & \small{{[}$\pm$8.0{]}}    & \small{{[}$\pm$13.2{]}} &  & \small{{[}$\pm$14.6{]}} & \small{{[}$\pm$7.0{]}}   & \small{{[}$\pm$9.0{]}}  &  & \small{{[}$\pm$15.7{]}} & \small{{[}$\pm$6.9{]}}   & \small{{[}$\pm$10.6{]}} \\
\multirow{2}{*}{$\Delta$F1 vs. best}    & -17.0        & \textbf{-10.4} & -13.0        &  & -9.3         & \textbf{-7.4} & -19.0        &  & -20.1        & \textbf{-5.1} & -10.0         \\
                         & \small{{[}$\pm$19.7{]}} & \small{{[}$\pm$13.3{]}}   & \small{{[}$\pm$14.8{]}} &  & \small{{[}$\pm$16.0{]}} & \small{{[}$\pm$7.3{]}}   & \small{{[}$\pm$11.6{]}} &  & \small{{[}$\pm$21.5{]}} & \small{{[}$\pm$5.0{]}}   & \small{{[}$\pm$8.2{]}} \\ \bottomrule
\end{tabular}
}
\end{table}
\clearpage 
Majority voting yields the most consistent F1 score improvements, generally outperforming the worst mapping direction, i.e., the lowest F1 score across $N$ possibilities, and reaching performance near the average. This behavior follows from its robustness to outliers, that is, poorly converged mapping directions or unusually well performing directions, which occur across all methods and scarcity scenarios (all source-to-target scores are shown in Table \ref{tab:fullresultsf1}). In contrast, the consensus and minority vote strategies produce noisy, inconsistent, and comparably worse improvements. We therefore restrict our following model comparisons to results obtained from applying the majority voting strategy. 

\subsection{Fault detection performance}
\begin{figure}[htb!]
    \centering
    \includegraphics[width=.9\textwidth]{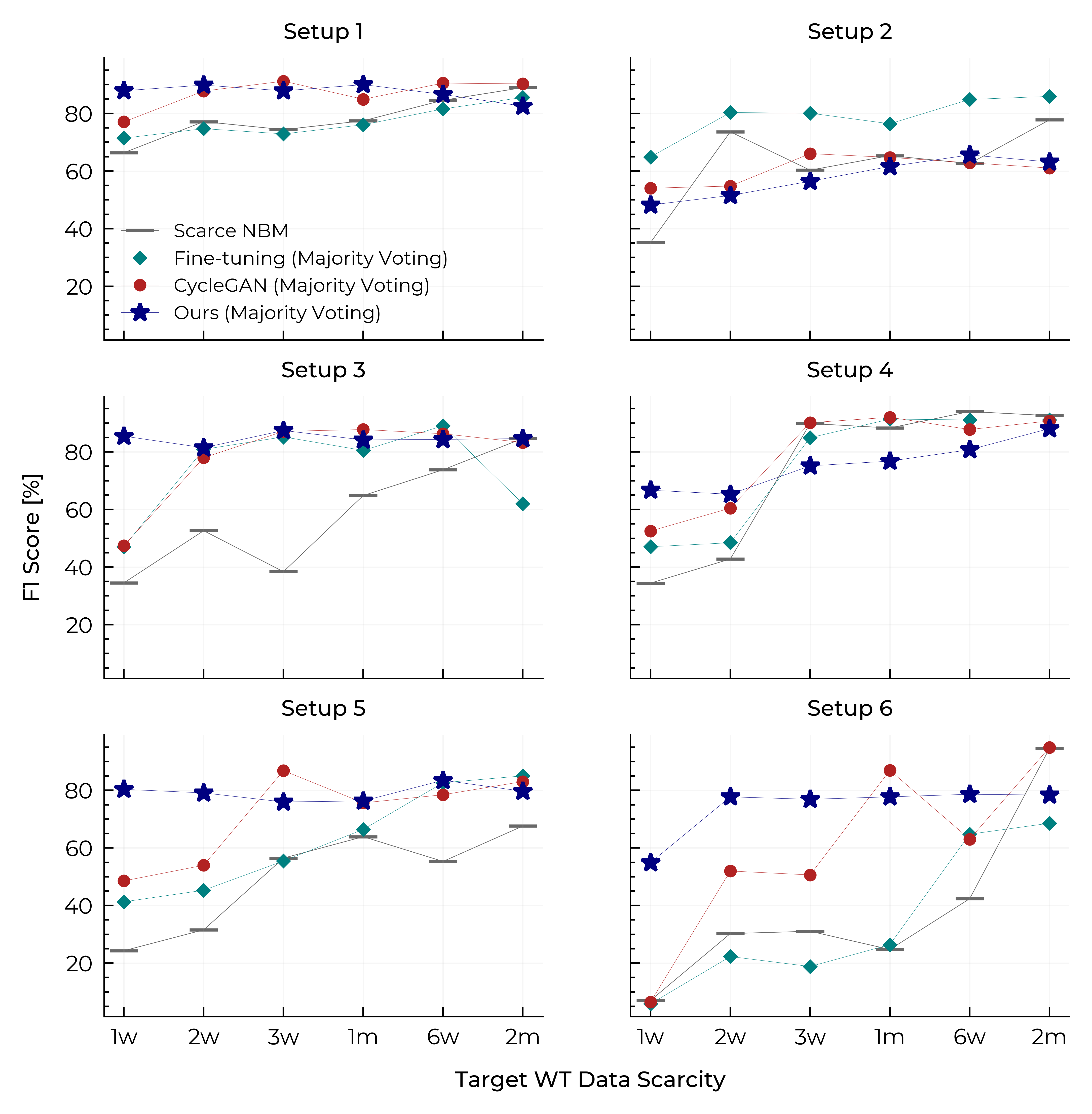}
    \caption{F1 scores showing the similarity of model scores to the ground truth for all setups and data scarcity scenarios.}
    \label{fig:setup_results}
\end{figure}

We visualize all F1 scores across setups and data scarcity scenarios from our proposed StarGAN-based model compared to the presented baselines in Figure \ref{fig:setup_results}. Detailed results are available in Table \ref{tab:f1-mv-results}.

\paragraph{Multi-source adapted baselines.}
The fault detection performance of the data-scarce NBMs is generally shown to significantly decrease with growing data scarcity. Training NBMs on only few weeks of collected training data can yield low anomaly score similarities compared to training models on representative training data. In a single-source setting, both fine-tuning and the CycleGAN-based single WT-to-WT domain mapping were found to achieve on average an improvement over data-scarce NBMs when training data is particularly scarce, underscoring the strength of domain adaptation for this task \cite{jonasFaultDetectionNew2025}. In our study, we adapted these baselines to a multi-source setting using ensemble fusion. Our results corroborate these previous findings and show that both domain adaptation methods can lead to improved fault detection, especially when training data is very scarce. In particular, the CycleGAN-based single-source WT-to-WT mapping managed to widely outperform both the data-scarce NBM and fine-tuning when considering scarcity scenarios with 1 month or less training data. In these scenarios, the resulting F1 scores were higher than those of the data-scarce NBM in 22/24 cases (91.7\%), while also outperforming fine-tuning in 19/24 cases (79.2\%). However, negative transfer learning persists as a possible outcome. We observe a decrease in relative performance especially for large training sets spanning 2 months, with the data-scarce NBM outperforming fine-tuning in 4/6 (66.7\%) and single-source domain mapping in 3/6 (50\%) setups, respectively. The transition to several available source domains cannot overcome this limitation, as its only change, the majority voting strategy, can only mitigate poorly trained directions. These findings suggest that single WT-to-WT domain mapping preserving SCADA states remains a viable and promising strategy in a multi-source setting to overcome poor fault detection performance in strong data scarcity scenarios. 

\paragraph{StarGAN-based mapping under severe data scarcity.}
We compare the adapted single-source baselines to our StarGAN-based multi-source domain mapping approach. We find that our model manages to largely outperform the data-scarce NBMs, fine-tuning, and CycleGAN-based single WT-to-WT mapping under severe data scarcity scenarios. With only one to two weeks of training data available, our model beats all baselines in 10 out of 12 (83.3\%) cases. For training datasets comprising one week of samples, our proposed method leads to a substantial average increase in F1 scores of +36.9\%, +24.3\%, and +22.9\% over the data-scarce NBM, fine-tuning, and CycleGAN-based domain mapping, respectively. In the scenario with two weeks of available training data, the performance improvements are +22.8\%, +15.5\%, and +9.7\% over the respective baselines.  

We attribute these improvements to the shared StarGAN generator learning translations across multiple data-rich source domains. The StarGAN model thereby learns shared representations from data-rich turbines that provide a helpful basis for out-of-distribution generalization of unseen target WT samples beyond the one or two weeks of available target domain samples. By mapping across several source domains, the model learns general patterns of normal behavior, reducing its dependence on the few available target samples. This prevent strong overfitting on scarce samples, furthermore providing training stability. In contrast, fine-tuning and CycleGAN-based domain mapping are constrained to learn exclusively from very limited observed modes and samples, resulting in comparably worse out-of-distribution performance. The advantage arising from using multiple source domains is supported by the CycleGAN-based domain mapping baseline. It largely shares the same objectives and architectural design (excluding required multi-source framework adjustments and capacity changes), and the ensemble fusion strategy. Yet, under severe data scarcity, our model results in more similar anomaly scores to the ground truth, suggesting that the primary benefit is the learning of source-to-source mappings.

Setup 2 is an exception, in which our model is outperformed by both single-source baselines in the 1- and 2 week scenarios. However, this setup behaves atypically overall, as the single-source domain mapping also exhibits poor performance compared to fine-tuning and even the data-scarce NBM. This may indicate issues in the training data or altered test performance behavior due to its comparably very high share of test incidents (see Table \ref{tab:datatable}), which may inadvertently assign high similarity scores to models biased towards producing high anomaly scores. More setups and further test sets are required to investigate potential failure cases.  

\paragraph{Performance with more available training data.}
Compared to the data-scarce NBM baseline, our proposed model continues to offer higher fault detection performance in data scarcity scenarios with 3 or more weeks of data, except for 2 months. Our model outperforms the baseline without transfer learning in 13/18 (72.2\%) of these cases. A relative performance decrease can especially be observed in setups and scenarios where representative data may already be available, i.e., when the data-scarce NBM offers a high similarity to the ground truth, such as in setup 4 under most scenarios. This negative transfer can occur across all our domain adaptation baselines.

More notably, however, our multi-source approach does not retain its advantages in relation to the single-source baselines in scarcity scenarios with more data (3 weeks to 2 months). Our StarGAN-based approach exceeds the F1 scores of both domain adaptation baselines in 7/24 cases (29.2\%). In the 1-month scenario (1m), our model results in an increase of +22.8\% and +8.2\% in F1 scores compared to the data-scarce NBM and fine-tuning, respectively, but ceases to outperform single-source domain mapping at -4.2\%. At 2 months of training data (2m), our model consistently results in slightly lower scores (-3.5\%, -0.3\%, -4.5\%, respectively). We observe that the relative benefit of shared representations diminishes as the single-source domain adaptation methods are exposed to more target samples. Once more training data becomes available, both single-source baselines can fit the models better on target data. This can be seen as a commonly steeper increase of performance over time compared to the more scenario-stable performance of our model.

\subsection{On unsupervised model selection}
\paragraph{Experiments.} 
We validated our proposed model selection proxy by training various model candidates and comparing the test set performance to our proxy metric $\Delta_T$. Model candidates share the same architecture and optimization configuration but differ in loss weights $\{\lambda_{cyc}, \lambda_{cls_D}, \lambda_{cls_G}, \lambda_0, \lambda_{\mathcal{R}}\}$. We evaluated varying candidates for all setups under different target WT data scarcities.

\begin{figure}[ht!]
    \centering
    \includegraphics[width=0.65\textwidth]{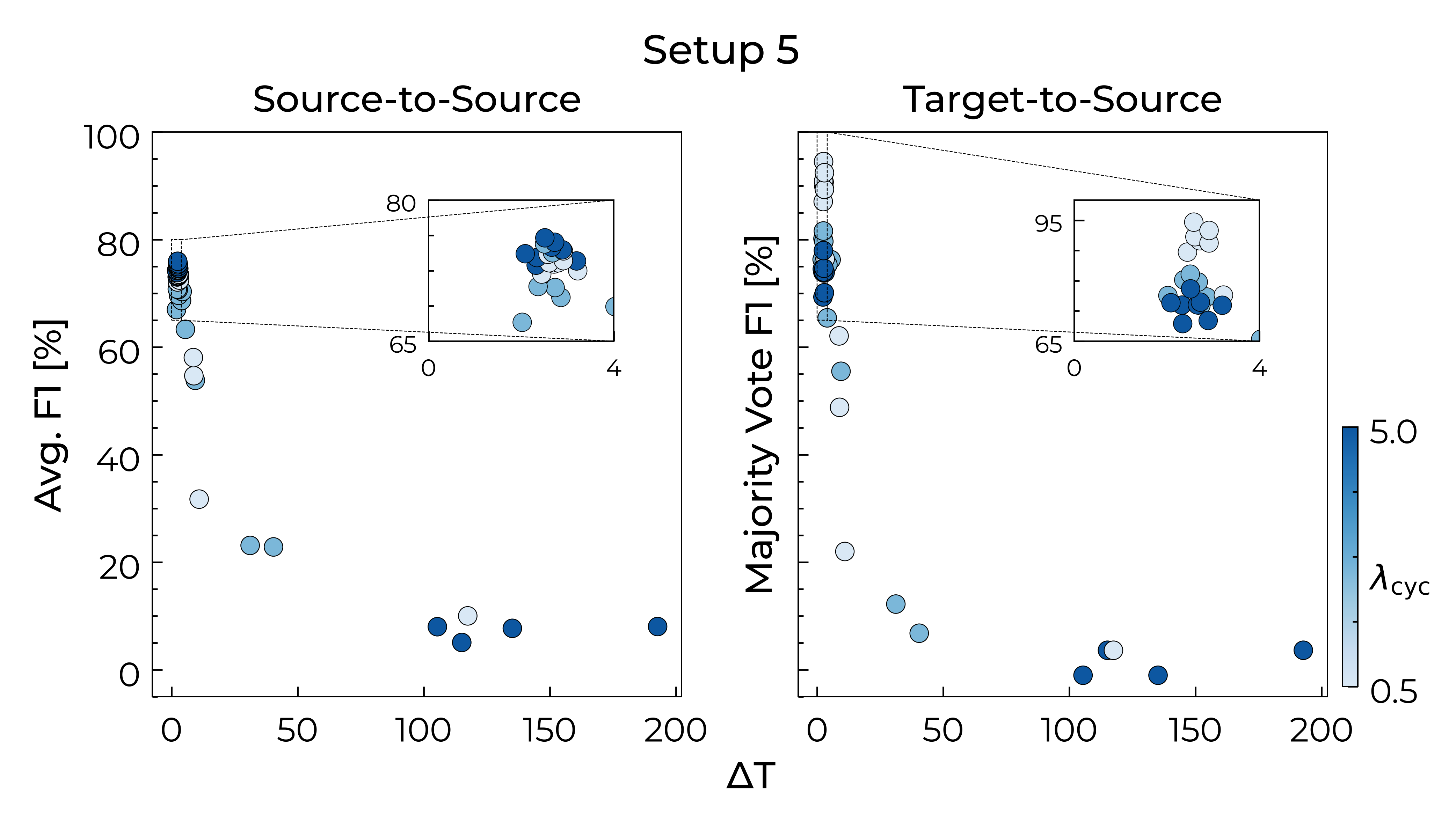}
    \caption{$\Delta_T$ of 36 different model candidates in comparison to the average F1 score when mapping data-rich source WTs to each other (left) and the F1 score with majority vote ensemble fusion when mapping the target WT to all source WTs. Results are shown for the model selection performed for setup 5 under 1 month of target WT data scarcity.}
    \label{fig:scatter}
\end{figure}

\paragraph{Results.} 
In Figure \ref{fig:scatter}, we first show the relationship of our proxy metric $\Delta_T$ to the average source-to-source mapping F1 score, that is, to how well the proxy metric correlates to the average fault detection performance of a data-rich test set mapped to other source WTs. We also show the relationship of $\Delta_T$ to the F1 score from mapping the data-scarce target test data to all source WTs. 

We observe that loss weights can substantially affect the model performance, as seen in the wide range of performance across the $36$ candidates. This further highlights the importance of model assessment at training time. Our findings show that our proxy metric can support the choice of loss weights largely by dismissing model candidates with very low F1 scores. Overall, a lower $\Delta_T$ is shown to correspond to higher F1 scores. This strong correlation is however largely driven by the separation of low- and high-performing candidates. A subset of low-performing model candidates exhibits very high $\Delta_T$ values with very low F1 scores, while a cluster of high-performing candidates with high F1 scores achieves very low $\Delta_T$ values. When restricting the analysis to high-performing candidates, as within the visualized $65-80\%$ and $65-95\%$ ranges, the correlation starts to diminish. Hence, our $\Delta_T$ metric is rather suitable as a filter to distinguish between low- and high-performing models, but offers little discriminative power to differentiate between high-performing models. We note that this filtering function is not unique to our metric, as the early stopping criterion also shows distinct values. Nonetheless, our process of selecting the model configuration based on the lowest $\Delta_T$ resulted in adequate model selection. While our proxy metric offers a promising direction towards unsupervised model selection, this may not necessarily hold in general and requires further research.  

We also notice that the F1 scores of high-performing models can vary more strongly for the target WT than for the averaged source WTs. On further investigation, we found that particularly $\lambda_{cyc}$ exhibits a strong turbine-specific effect. An increased weight of the cycle-consistency loss led to improvements in some turbines but degraded the performance of others. These changes can offset each other across several source turbines, causing similar metric and performance values. However, the per-turbine sensitivity becomes more impactful for the single target WT. Certain $\lambda_{cyc}$ choices led to excellent downstream fault detection (e.g., F1 scores larger than 90\%), while other weights yielded less reliable fault detection (e.g., 65\%). Therefore, a very strong performance could be attainable for the data-scarce target turbine, but cannot be selected by our metric only considering mappings across source WTs. Consequently, our proposed metric cannot be employed to reliably return the best performing model specifically for the target WT but is shown to distinguish between low- and high-performing models.

\subsection{Mapping example}
\begin{figure}[htb!]
    \centering
    \includegraphics[width=.55\textwidth]{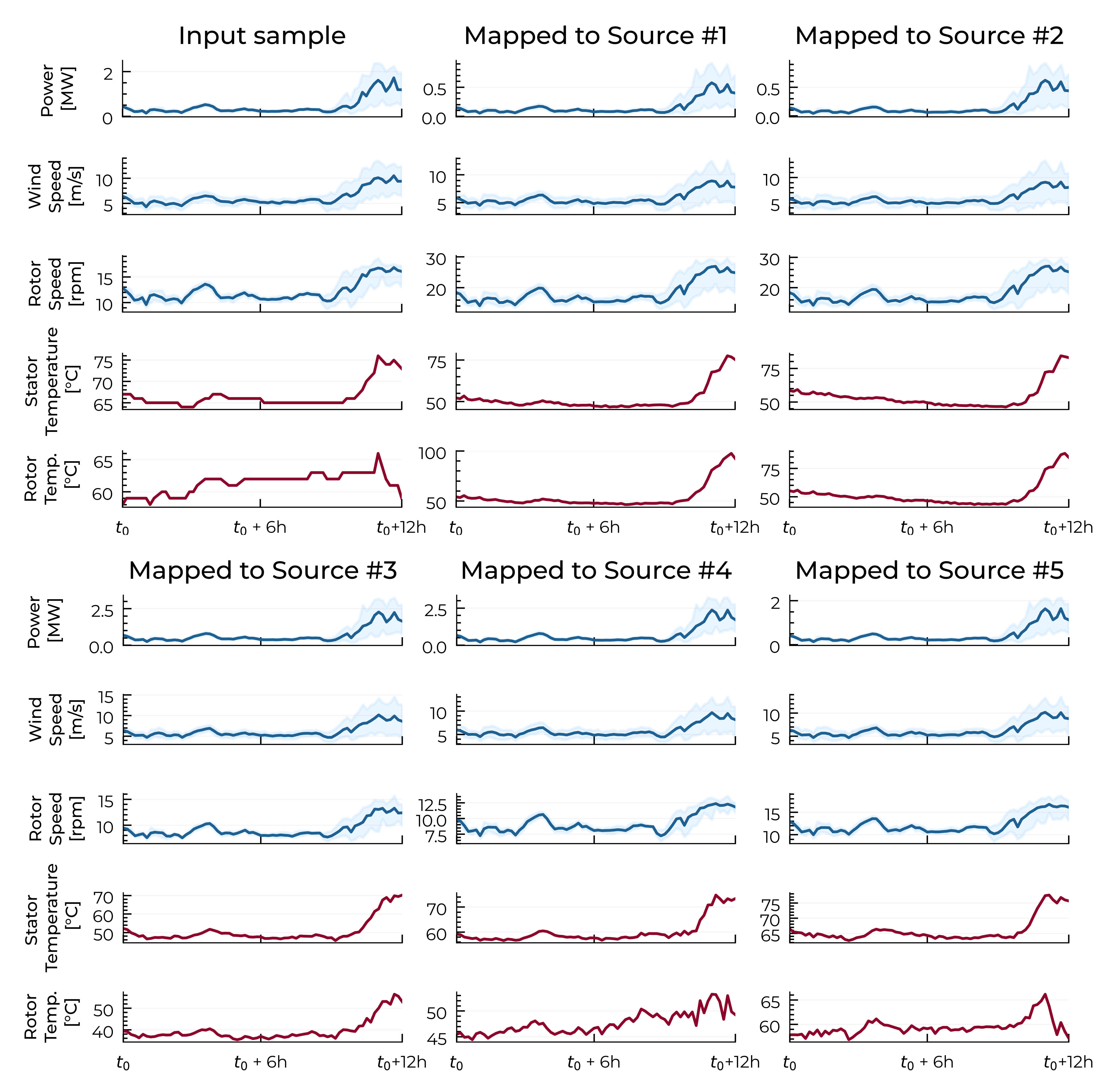}
    \caption{An input sample of target WT05 mapped to each source domains (setup 5). The model was trained on 3 weeks of target data.}
    \label{fig:mappedsamples}
\end{figure}

\begin{figure}[htb!]
    \centering
    \includegraphics[width=0.45\textwidth]{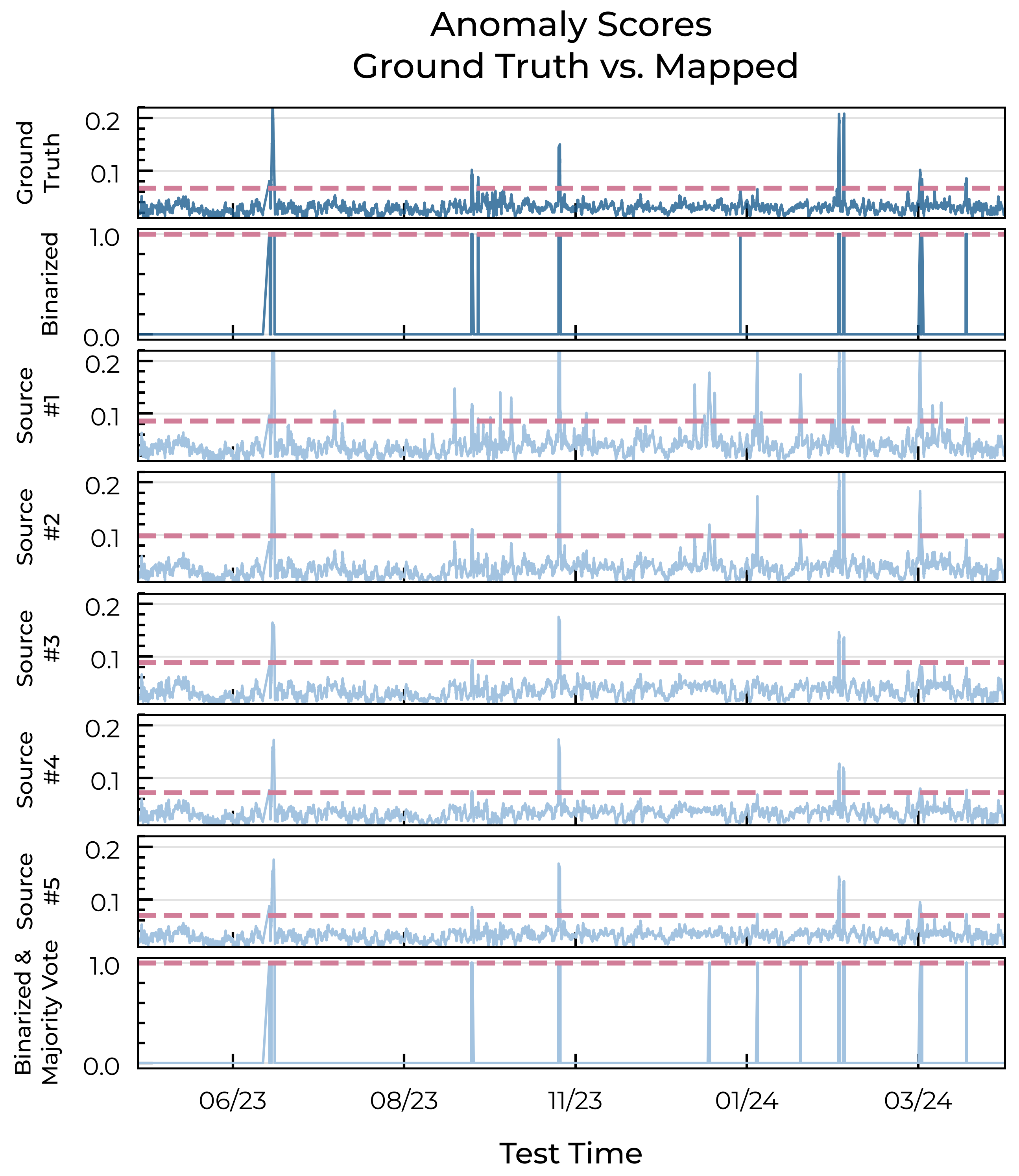}
    \caption{Anomaly scores for test data of target WT05 (setup 5). The ground truth (top) is compared samples mapped to and scored in each source domain. The model was trained on 3 weeks of target data.}
    \label{fig:ascores}
\end{figure}

In Figure \ref{fig:mappedsamples}, an example of an input sample mapped to several source domains is shown. The preservation of the original operation states is clearly visible, with the model adjusting for scaling differences and minor modifications in dynamics across WTs. Figure \ref{fig:ascores} shows a comparison of ground truth anomaly scores, binarized scores, evaluated mapped samples, and the binarized and fused final scores. 
\vspace{-0.5cm}

\subsection{Limitations and future research directions}

\paragraph{StarGAN-based multi-domain WT mapping.}
Our multi-source framework is shown to be particularly effective under strong data scarcity. In scenarios with only one to two weeks of accumulated training data, it can substantially improve fault detection, further underlining the effectiveness of domain mapping of WTs with limited operating history, such as newly commissioned wind turbines. However, there are notable limitations. The strong variance in F1 scores across mapping directions and setup compositions indicates that the choice of source turbines has a considerable effect on performance. Moreover, the representativeness of the limited training sets, and thus the performance of the domain adaptation framework, depends not only on the number of weeks but also on contained operational states. Therefore, selecting an appropriate method at training and deployment time remains difficult, since neither achieved performance nor the relative relative ranking can be determined. As such, we cannot identify the best performing strategy in advance.  

Future research should further investigate the effects on performance depending on the number and types of source turbines. Furthermore, studies on datasets with annotated faults would allow an assessment of whether specific fault types affect domain mapping differently. Our approach should also be validated on different SCADA systems and variables with WTs from more manufacturers. Lastly, there may be potential for improvements in efficiency, such as continual learning to update the shared StarGAN generator to newly arrived data instead of retraining, or whether a pretrained source-exclusive StarGAN model can be efficiently adapted to small target turbine datasets. 

\paragraph{Unsupervised model selection.} 
While we show that our proposed proxy metric $\Delta_T$ can support model selection, it carries several limitations. First, rather than predicting final performance, our metric mainly separates poor from functional mappings. Thus, it depends on a sufficiently broad set of model candidates to ensure that reliable models are included. Moreover, the absence of anomalies restricts any assessment or tuning of techniques that explicitly only affect the mapping of anomalies, such as our anomaly augmentation step or certain architecture choices. Assessing and tuning such components therefore still requires validation on real anomalies. Lastly, models that do not preserve the operational state of the input sample by setting $\lambda_{cyc}, \lambda_{0}$, $\lambda_{\mathcal{R}}$ to 0 can nonetheless yield a low $\Delta_T$, as unconstrained GANs also learn to generate samples yielding realistic anomaly scores. Detailed checks or visual inspections may be helpful to filter out non-preserving models. Lastly, an observed dependence of performance on $\lambda_{cyc}$ further suggests potential for adapting the cycle-consistency loss, for example, through turbine-specific or dynamic weighting schemes or through alternative consistency formulations. Given these limitations, we propose this metric largely as a helpful filter to exclude suboptimal model configurations. Reliable anomaly-free selection of the best model candidate remains a very important open challenge.


\section{Conclusions}
In our study, we investigated multi-source domain adaptation for wind turbine fault detection under training data scarcity. Normal behavior models can produce unreliable anomaly scores when only limited training data is available, since they require representative samples of the turbine's normal operation. To this end, we introduce a generative domain mapping model that translates SCADA measurements of a data-scarce target turbine to resemble measurements of multiple data-rich source turbines. As opposed to previous single-source domain mapping approaches, our StarGAN-based model maps samples from a data-scarce target turbine to several data-rich source turbines simultaneously. Our model preserves the operational state of the input SCADA samples during mapping such that, for instance, idle states or anomalous states are preserved during translation. The mapped operational data can thus be evaluated by pretrained and reliable normal behavior models of the source turbines.

We validated our approach with field data of 7 operational wind turbines arranged into various target- and source domain configurations. The target turbine was affected by a data scarcity limiting its training samples to 1-8 weeks. We found that our model largely outperformed single-source fine-tuning and domain mapping for fault detection under severe data scarcity scenarios in which only 1-2 weeks of training data were available. We attributed this observation to our framework learning shared representations from mapping across multiple source domains. However, this no longer holds as target training data becomes more available. Lastly, we addressed the challenge of assessing operational performance at training time without anomalies, as we show a substantial effect of hyperparameters on fault detection performance. As a step towards unsupervised model selection, we proposed a proxy metric computed entirely on fault-free data. While our proxy metric cannot sufficiently discriminate among well-performing model candidates, it can reliably filter out poorly performing model configurations. Overall, our results suggest that multi-source domain mapping shows promising potential to improve fault detection for turbines that strongly lack training data. Future work could extend the validation of the proposed model to additional datasets and wind turbines from different manufacturers. We outline several promising research directions, including investigating the role of source domains, developing more efficient training and adaptation schemes, and improving fault-free model selection strategies.

\section*{Acknowledgements}
\vspace{-0.3cm}
This research was funded by the Swiss National Science Foundation under grant number 206342. We want to thank aventron AG (Weidenstrasse 27, 4142 Münchenstein, Switzerland, \url{www.aventron.com}) for sharing their measurement data enabling this research.
\vspace{-0.2cm}
\printbibliography

\setcounter{table}{0}
\renewcommand{\thetable}{A\arabic{table}}

\setcounter{figure}{0}
\renewcommand{\thefigure}{A\arabic{figure}}

\section*{Appendix}
\section*{Appendix A: Dataset Overview}
\begin{table}[ht!]
\centering
\caption{Details of all 7 WTs used in our study. Table adapted from \cite{jonasFaultDetectionNew2025}, which used the same dataset. The stated rated power value is determined based on the observed data and used for the filtering process and the rated power loss.}
\label{tab:datatable}
\resizebox{.8\textwidth}{!}{%
\begin{tabular}{@{}llllllll@{}}
\toprule
&
  \multicolumn{1}{c}{WT01} &
  \multicolumn{1}{c}{WT02} &
  \multicolumn{1}{c}{WT03} &
  \multicolumn{1}{c}{WT04} &
  \multicolumn{1}{c}{WT05} &
  \multicolumn{1}{c}{WT06} &
  \multicolumn{1}{c}{WT07} \\ \midrule
\textbf{Specifications}                                                              &         &         &         &         &         &         &         \\
Location                                                                      & Onshore & Onshore & Onshore & Onshore & Onshore & Onshore & Onshore \\
Set Rated Power {[}KW{]}                                                          & 800     & 3000    & 2350    & 2050    & 2300    & 800     & 3050    \\
                                                                              &         &         &         &         &         &         &         \\
\textbf{Training and Validation Set}                                                         &         &         &         &         &         &         &         \\
\begin{tabular}[c]{@{}l@{}}Days between \\ first and last sample\end{tabular} & 899     & 877     & 706     & 831     & 827     & 798     & 830     \\
\multirow{2}{*}{\begin{tabular}[c]{@{}l@{}}Filtered\\ 12h SCADA samples\end{tabular}} &
  \multirow{2}{*}{50574} &
  \multirow{2}{*}{58403} &
  \multirow{2}{*}{66407} &
  \multirow{2}{*}{70191} &
  \multirow{2}{*}{70922} &
  \multirow{2}{*}{70197} &
  \multirow{2}{*}{66117} \\
                                                                              &         &         &         &         &         &         &         \\
                                                                              &         &         &         &         &         &         &         \\
\textbf{Test Set}                                                             &         &         &         &         &         &         &         \\
\begin{tabular}[c]{@{}l@{}}Days between\\ first and last sample\end{tabular}  & 387     & 387     & 323     & 355     & 354     & 341     & 355     \\
\multirow{2}{*}{\begin{tabular}[c]{@{}l@{}}Unfiltered \\ 12h-samples\end{tabular}} &
  \multirow{2}{*}{48736} &
  \multirow{2}{*}{45593} &
  \multirow{2}{*}{43309} &
  \multirow{2}{*}{35573} &
  \multirow{2}{*}{37308} &
  \multirow{2}{*}{21187} &
  \multirow{2}{*}{43107} \\
                                                                              &         &         &         &         &         &         &         \\
\multicolumn{1}{r}{samples with incidents {[}\%{]}} &
  \multicolumn{1}{r}{18.6} &
  \multicolumn{1}{r}{43.1} &
  \multicolumn{1}{r}{23.3} &
  \multicolumn{1}{r}{9.2} &
  \multicolumn{1}{r}{2.1} &
  \multicolumn{1}{r}{0.6} &
  \multicolumn{1}{r}{18.0} \\ \bottomrule
\end{tabular}%
}
\end{table}

\section*{Appendix B: Training Details}
\label{sec:AppendixB}
\paragraph{Normal behavior models.} For each WT, an autoencoder-based NBM was trained with representative data (full training set) and for all data scarcity scenarios (data-scarce NBM baseline; 1 to 8 weeks). The models were trained by minimizing the mean squared reconstruction error of the WT's fault-free 12h-SCADA samples. Optimization was done with the RAdam \cite{liu2019radam} optimizer at a learning rate of $0.003$ and a batch size of $128$. Early stopping was triggered when the reconstruction error of normal validation data stopped decreasing. The model architecture for the autoencoder-based normal behavior model is described in Table \ref{tab:NBM}. The NBMs are identical to those in \cite{jonasFaultDetectionNew2025}. We reused the pretrained models directly, matching the reported NBM results from our previous study without modification. 

\begin{table}[htp!]
\centering
\caption{Architecture of the normal behavior models used in our study.}
\label{tab:NBM}
\resizebox{0.465\textwidth}{!}{%
\begin{tabular}{@{}lll@{}}
\toprule
\multicolumn{3}{c}{\textbf{Autoencoder-based NBM Architecture}}                     \\ \midrule
\multicolumn{3}{c}{{\ul \textit{Input}}}                                            \\
\multicolumn{3}{l}{11 channels x 72 datapoints}                                     \\
\multicolumn{3}{c}{{\ul \textit{Encoder}}}                                          \\
\multicolumn{3}{l}{1D Convolution (32 filters, kernel size 7, stride 1, Mish \cite{misraMishSelfRegularized2020})  x 2} \\
\multicolumn{3}{l}{MaxPool (kernel size 2)}                                         \\
\multicolumn{3}{l}{GroupNorm   (1 group, 32 channels)}                              \\
\multicolumn{3}{l}{1D Convolution (32 filters, kernel size 5, stride 1, Mish)  x 2} \\
\multicolumn{3}{l}{MaxPool (kernel size 2)}                                         \\
\multicolumn{3}{l}{GroupNorm   (1 group, 32 channels)}                              \\
\multicolumn{3}{l}{Flatten;  FC(32 x 18, 72)}                                       \\
\multicolumn{3}{c}{{\ul \textit{Decoder}}}                                          \\
\multicolumn{3}{l}{Upsample (factor 2)}                                             \\
\multicolumn{3}{l}{1D Convolution (32 filters, kernel size 5, stride 1, Mish)  x 2} \\
\multicolumn{3}{l}{GroupNorm   (1 group, 32 channels)}                              \\
\multicolumn{3}{l}{Upsample (factor 2)}                                             \\
\multicolumn{3}{l}{1D Convolution (32 filters, kernel size 5, stride 1, Mish)  x 2} \\
\multicolumn{3}{l}{GroupNorm   (1 group, 32 channels)}                              \\
\multicolumn{3}{l}{1D Convolution (11 dilters, kernel size 1, stride 1, bias, linear)} \\ \bottomrule
\end{tabular}%
}
\end{table}

\paragraph{StarGAN-based domain mapping model.} 
The architecture of our StarGAN is described in Table \ref{tab:stargan_arch}. The TCN-based generator and the 1D convolutional discriminator follow \cite{jonasFaultDetectionNew2025}, which we adapted to the multi-source setting by increasing model capacity and adjusting the inputs and outputs to accommodate for the destination label and classification. We trained our StarGAN-based model in iterations described in Algorithm \ref{alg:1}. Adam ($\beta_1 = 0.5, \beta_2 = 0.999$) was used for optimization with a learning rate of $0.0001$ and a batch size of $32$. Instead of using the final training weights, we use an exponential moving average (EMA) of the generator weights at inference time, a technique that has been shown to improve GAN training \cite{gan_ema_paper}.

\begin{table}[htp!]
\centering
\caption{Left: Generator architecture. The residual block structure is illustrated in Figure \ref{fig:resblock}. Right: Discriminator architecture.}
\label{tab:stargan_arch}
\resizebox{0.95\textwidth}{!}{%
\begin{tabular}{@{}cllllcll@{}}
\cmidrule(r){1-3} \cmidrule(l){6-8}
\multicolumn{3}{c}{\textbf{Generator Architecture}} &
   &
   &
  \multicolumn{3}{c}{\textbf{Discriminator Architecture}} \\ \cmidrule(r){1-3} \cmidrule(l){6-8} 
\multicolumn{1}{l}{} &
   &
   &
   &
   &
  \multicolumn{1}{l}{} &
   &
   \\
\multicolumn{3}{c}{{\ul \textit{Inputs}}} &
   &
   &
  \multicolumn{3}{c}{{\ul \textit{Input}}} \\
\multicolumn{1}{l}{{\ul \textit{SCADA Input}}} &
   &
  \multicolumn{1}{r}{{\ul \textit{Destination Label}}} &
   &
   &
  \multicolumn{3}{c}{11 channels   x 72 datapoints} \\
\multicolumn{1}{l}{11 channels x 72 datapoints} &
   &
  \multicolumn{1}{r}{N+1 channels} &
   &
   &
  \multicolumn{1}{l}{} &
   &
   \\
\multicolumn{3}{c}{Concat (11 + (N+1) channels x 72 datapoints)} &
   &
   &
  \multicolumn{3}{c}{{\ul \textit{Block 1}}} \\
\multicolumn{3}{c}{\textit{\textbf{}}} &
   &
   &
  \multicolumn{3}{c}{1D Convolution (256 filters, kernel size 5, stride 2, Mish)} \\
\multicolumn{3}{c}{{\ul \textit{TCN Blocks}}} &
   &
   &
  \multicolumn{3}{c}{GroupNorm   (1 group, 256 channels)} \\
\multicolumn{3}{c}{TCN-Residual Block (64, 3, 1, False)} &
   &
   &
  \multicolumn{3}{l}{} \\
\multicolumn{3}{c}{TCN-Residual Block (64, 3, 2, True)} &
   &
   &
  \multicolumn{3}{c}{{\ul \textit{Block 2}}} \\
\multicolumn{3}{c}{TCN-Residual Block (64, 3, 4, True)} &
   &
   &
  \multicolumn{3}{c}{1D Convolution (256 filters, kernel size 3, stride 2, Mish)} \\
\multicolumn{3}{c}{TCN-Residual Block   (64, 3, 8, True)} &
   &
   &
  \multicolumn{3}{c}{GroupNorm   (1 group, 256 channels)} \\
\multicolumn{3}{c}{TCN-Residual Block   (64, 3, 16, True)} &
   &
   &
  \multicolumn{3}{l}{} \\
\multicolumn{3}{c}{TCN-Residual Block (64, 3, 32, False)} &
   &
   &
  \multicolumn{3}{c}{{\ul \textit{Block 3}}} \\
\multicolumn{3}{c}{\textit{\textbf{}}} &
   &
   &
  \multicolumn{3}{c}{1D Convolution (512 filters, kernel size 3, stride 2, Mish)} \\
\multicolumn{3}{c}{{\ul \textit{Output Layer}}} &
   &
   &
  \multicolumn{3}{c}{GroupNorm   (1 group, 512 channels)} \\
\multicolumn{3}{c}{1D Convolution (11 filters, kernel size 1, stride 1, bias, linear)} &
   &
   &
  \multicolumn{3}{c}{Flatten} \\
\multicolumn{1}{l}{} &
   &
   &
   &
   &
  \multicolumn{3}{l}{} \\
\multicolumn{1}{l}{} &
   &
   &
   &
   &
  \multicolumn{3}{c}{{\ul \textit{Output Layers}}} \\
\multicolumn{1}{l}{} &
   &
   &
   &
   &
  \multicolumn{1}{l}{{\ul \textit{Critic}}} &
   &
  \multicolumn{1}{r}{{\ul \textit{Classifier}}} \\
\multicolumn{1}{l}{} &
   &
   &
   &
   &
  \multicolumn{2}{l}{FC (9 x 512,   1)} &
  \multicolumn{1}{r}{FC(9 x 512, N+1)} \\ \cmidrule(r){1-3} \cmidrule(l){6-8} 
\end{tabular}%
}
\end{table}

\begin{figure}[htb!]
    \centering
    \includegraphics[width=.25\textwidth]{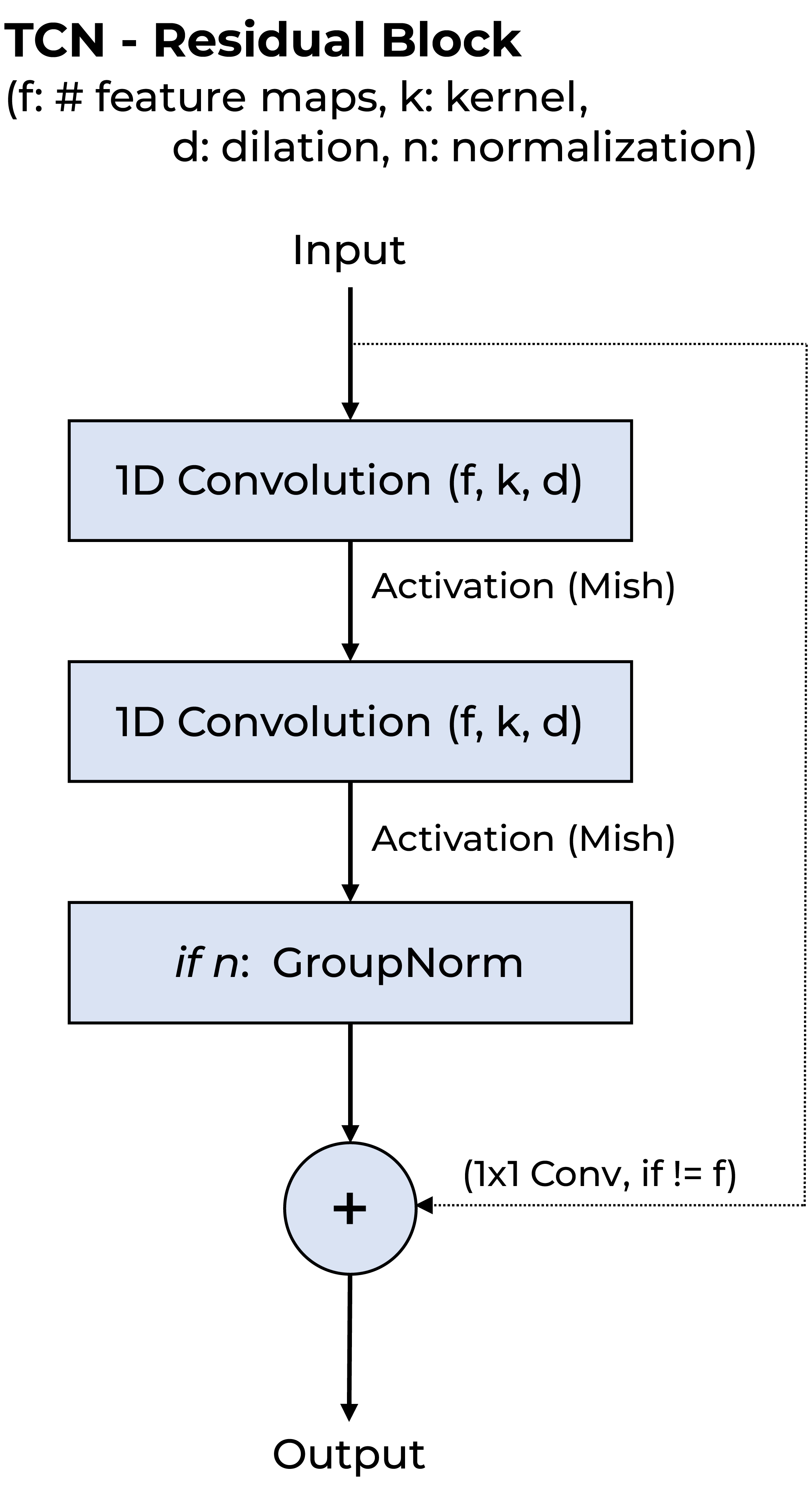}
    \caption{Residual block architecture, as used in \cite{jonasFaultDetectionNew2025}.}
    \label{fig:resblock}
\end{figure}

\begin{algorithm}[ht!]
\caption{Our StarGAN-based multi-source WT mapping algorithm.}
\label{alg:1}
\begin{algorithmic}[1]
\Require{StarGAN generator $G$ and discriminator $D$, loss weight hyperparameters $\{\lambda_{cyc}, \lambda_{cls_D}, \lambda_{cls_G}, \lambda_0, \lambda_{\mathcal{R}}\}$.}
\While{early stopping is not triggered}
\For{each origin domain}

    \State Sample a batch $b_o$ from the origin domain
    \State Sample a batch $b_d$ from a different, randomly selected destination domain

    \textbf{Generator updates}
    \State Map batches across domains: 
    $b_{od} = G(b_o, c_d), b_{do}=G(b_d, c_o)$
    \State Cycle batches back to their origin domain: $b_{odo} = G(b_{od}, c_o), b_{dod}=G(b_{do}, c_d)$
    \State $\mathcal{L}_{GAN_{QP}} \leftarrow \mathcal{L}_{GAN_{QP}}(b_o, b_{do}) +  \mathcal{L}_{GAN_{QP}}(b_d, b_{od})$  \# both mapping directions 
    \State $\mathcal{L}^f_{cls} \leftarrow \mathcal{L}^f_{cls}(b_{do}, c_o) +  \mathcal{L}^f_{cls}(b_{od}, c_d)$ \# classification loss from $D$ for \textit{mapped} samples 
    \State $\mathcal{L}_{G} \leftarrow \mathcal{L}_{GAN_{QP}} + \lambda_{{cls}_G}\mathcal{L}^r_{cls} + \lambda_{cyc} \mathcal{L}_{cyc} + \lambda_{0} \mathcal{L}_0 + \lambda_R \mathcal{L}_R$
    \State Artificially corrupt batches $b_o$ and $b_d$;  map back and forth (cycle) and add $\lambda_{cyc} \mathcal{L}_{cyc}$ to $\mathcal{L}_{G}$
    \State Update $\theta_G$ using Adam on $\nabla_{\theta_G} \mathcal{L}_G$

    \textbf{Discriminator updates}
    \State $\mathcal{L}_{GAN_{{QP}_D}} \leftarrow\mathcal{L}_{GAN_{{QP}_D}}(b_o, b_{do}) + \mathcal{L}_{GAN_{{QP}_D}}$
    \State $\mathcal{L}^r_{cls} \leftarrow \mathcal{L}^r_{cls}(b_o, c_o) + \mathcal{L}^r_{cls}(b_d, c_d)$ \# classification loss of \textit{real} samples
    \State $\mathcal{L}_{D} \leftarrow \mathcal{L}_{GAN_{{QP}_D}} + \mathcal{L}^r_{cls}$
    \State Update $\theta_D$ using Adam on $\nabla_{\theta_D} \mathcal{L}_D$
\EndFor
\EndWhile
\end{algorithmic}
\end{algorithm}

\paragraph{GAN framework.} For the adversarial objective we employ the GAN with Quadratic Potential (GAN-QP) framework \cite{suGANQPNovelGAN2018}:
\begin{equation}\begin{aligned}&T= \mathop{\arg\max}_T\, \mathbb{E}_{(x_r,x_f)\sim p(x_r)q(x_f)}\left[T(x_r)-T(x_f) - \frac{(T(x_r)-T(x_f))^2}{2\lambda_{QP} d(x_r,x_f)}\right] \\
&G = \mathop{\arg\min}_G\,\mathbb{E}_{(x_r,x_f)\sim p(x_r)q(x_f)}\left[T(x_r)-T(x_f)\right]
\end{aligned}\end{equation}
$G$ denotes the generator and $T$ the discriminator, while $x_r$ and $x_f$ denote the real and fake (mapped) samples, respectively. The value of $\lambda_{QP}$ was set to $1$ and we use the $L_1$ norm as the distance metric $d$.

\paragraph{Implementation.} Our implementation is based on PyTorch and training was performed on a single NVIDIA GPU. We release our code publicly on GitHub \url{https://github.com/EnergyWeatherAI/Multi_Domain_WT_Mapping}.

\clearpage 
\section*{Appendix C: Results}
\begin{table}[htb!]
\centering
\caption{F1 Scores [\%] for all baselines, setups, and data scarcity scenarios. The F1 scores are obtained by applying a majority voting strategy to combine $N$ binarized anomaly scores. Within each setup and data scarcity scenario, the bold score represents the best value and the second best performance is underlined.}
\label{tab:f1-mv-results}
\resizebox{\textwidth}{!}{%
\begin{tabular}{@{}lcccclcccclcccc@{}}
\toprule
\multirow{2}{*}{\begin{tabular}[c]{@{}l@{}}Setup \\ (\# source WTs)\end{tabular}} &
  \multicolumn{4}{c}{2 months} &
   &
  \multicolumn{4}{c}{6 weeks} &
   &
  \multicolumn{4}{c}{1 month} \\ \cmidrule(l){2-15} 
 &
  Scarce &
  FT-MV &
  CG-MV &
  Ours-MV &
   &
  Scarce &
  FT-MV &
  CG-MV &
  Ours-MV &
   &
  Scarce &
  FT-MV &
  CG-MV &
  Ours-MV \\ \midrule
WT01 (4) &
  {\ul 89.0} &
  \multicolumn{1}{l}{85.6} &
  \textbf{90.3} &
  82.5 &
   &
  84.6 &
  \multicolumn{1}{l}{81.6} &
  \multicolumn{1}{l}{\textbf{90.5}} &
  \multicolumn{1}{l}{{\ul 86.6}} &
   &
  77.4 &
  76.1 &
  {\ul 84.9} &
  \textbf{90.0} \\
WT02 (4) &
  {\ul 77.8} &
  \multicolumn{1}{l}{\textbf{85.9}} &
  61.0 &
  63.1 &
   &
  62.6 &
  \multicolumn{1}{l}{\textbf{84.9}} &
  \multicolumn{1}{l}{62.9} &
  \multicolumn{1}{l}{{\ul 65.6}} &
   &
  {\ul 65.3} &
  \textbf{76.4} &
  64.7 &
  61.6 \\
WT03 (6) &
  \textbf{84.7} &
  \multicolumn{1}{l}{62.0} &
  83.3 &
  {\ul 84.6} &
   &
  {\ul 73.8} &
  \multicolumn{1}{l}{\textbf{89.1}} &
  \multicolumn{1}{l}{86.3} &
  \multicolumn{1}{l}{84.3} &
   &
  64.8 &
  80.5 &
  \textbf{87.8} &
  {\ul 84.1} \\
WT04 (5) &
  \textbf{92.6} &
  \multicolumn{1}{l}{{\ul 91.1}} &
  90.7 &
  88.1 &
   &
  \textbf{94.0} &
  \multicolumn{1}{l}{{\ul 91.1}} &
  \multicolumn{1}{l}{87.8} &
  \multicolumn{1}{l}{80.8} &
   &
  88.3 &
  {\ul 91.4} &
  \textbf{92.0} &
  76.8 \\
WT05 (5) &
  67.6 &
  \multicolumn{1}{l}{\textbf{85.0}} &
  {\ul 83.0} &
  79.7 &
   &
  55.3 &
  \multicolumn{1}{l}{{\ul 82.6}} &
  \multicolumn{1}{l}{78.5} &
  \multicolumn{1}{l}{\textbf{83.5}} &
   &
  63.9 &
  66.4 &
  {\ul 75.6} &
  \textbf{76.3} \\
WT06 (4) &
  {\ul 94.5} &
  \multicolumn{1}{l}{68.6} &
  \textbf{94.9} &
  78.3 &
   &
  42.4 &
  \multicolumn{1}{l}{{\ul 64.8}} &
  \multicolumn{1}{l}{63.0} &
  \multicolumn{1}{l}{\textbf{78.6}} &
   &
  24.7 &
  26.4 &
  \textbf{86.9} &
  {\ul 77.7} \\
 &
  \multicolumn{1}{l}{} &
  \multicolumn{1}{l}{} &
  \multicolumn{1}{l}{} &
  \multicolumn{1}{l}{} &
   &
  \multicolumn{1}{l}{} &
  \multicolumn{1}{l}{} &
  \multicolumn{1}{l}{} &
  \multicolumn{1}{l}{} &
   &
  \multicolumn{1}{l}{} &
  \multicolumn{1}{l}{} &
  \multicolumn{1}{l}{} &
  \multicolumn{1}{l}{} \\ \midrule
 &
  \multicolumn{4}{c}{3 weeks} &
  \multicolumn{1}{c}{} &
  \multicolumn{4}{c}{2 weeks} &
  \multicolumn{1}{c}{} &
  \multicolumn{4}{c}{1 week} \\ \cmidrule(lr){2-5} \cmidrule(lr){7-10} \cmidrule(l){12-15} 
 &
  Scarce &
  FT-MV &
  CG-MV &
  Ours-MV &
  \multicolumn{1}{c}{} &
  Scarce &
  FT-MV &
  CG-MV &
  Ours-MV &
  \multicolumn{1}{c}{} &
  Scarce &
  FT-MV &
  CG-MV &
  Ours-MV \\ \midrule
WT01 (4) &
  74.4 &
  72.9 &
  \textbf{91.2} &
  {\ul 87.9} &
  \multicolumn{1}{c}{{\ul }} &
  77.1 &
  74.7 &
  {\ul 87.8} &
  \textbf{89.8} &
  \multicolumn{1}{c}{{\ul }} &
  66.4 &
  71.4 &
  {\ul 77.1} &
  \textbf{87.9} \\
WT02 (4) &
  60.3 &
  \textbf{80.1} &
  {\ul 66.0} &
  56.4 &
  \multicolumn{1}{c}{} &
  {\ul 73.6} &
  \textbf{80.3} &
  54.8 &
  51.5 &
  \multicolumn{1}{c}{} &
  35.2 &
  \textbf{64.9} &
  {\ul 54.0} &
  48.2 \\
WT03 (6) &
  38.4 &
  85.2 &
  {\ul 87.1} &
  \textbf{87.4} &
  \multicolumn{1}{c}{} &
  52.6 &
  {\ul 80.9} &
  78.0 &
  \textbf{81.4} &
  \multicolumn{1}{c}{\textbf{}} &
  34.5 &
  47.1 &
  {\ul 47.4} &
  \textbf{85.5} \\
WT04 (5) &
  {\ul 89.8} &
  84.9 &
  \textbf{90.1} &
  75.1 &
  \multicolumn{1}{c}{} &
  42.8 &
  48.4 &
  60.5 &
  \textbf{65.3} &
  \multicolumn{1}{c}{\textbf{}} &
  34.4 &
  47.1 &
  {\ul 52.5} &
  \textbf{66.7} \\
WT05 (5) &
  56.5 &
  55.4 &
  \textbf{86.8} &
  {\ul 76.0} &
  \multicolumn{1}{c}{{\ul }} &
  31.5 &
  45.3 &
  {\ul 54.0} &
  \textbf{79.1} &
  \multicolumn{1}{c}{\textbf{}} &
  24.3 &
  41.3 &
  {\ul 48.5} &
  \textbf{80.3} \\
WT06 (4) &
  31.1 &
  18.8 &
  {\ul 50.6} &
  \textbf{76.9} &
  \multicolumn{1}{c}{\textbf{}} &
  30.3 &
  22.3 &
  {\ul 52.0} &
  \textbf{77.7} &
  \multicolumn{1}{c}{\textbf{}} &
  {\ul 7.0} &
  5.9 &
  6.5 &
  \textbf{54.9} \\ \bottomrule
\end{tabular}%
}
\end{table}
\clearpage
\section*{Appendix D: Detailed Results}
\begin{table}[htb!]
\centering
\caption{Detailed F1 Scores [\%] for all target-to-source mappings across all setups and data scarcity scenarios. FT: Fine-tuning. CG: CycleGAN-based single WT-to-WT mapping. Ours: StarGAN-based WT-to-WT mapping.}
\label{tab:fullresultsf1}
\resizebox{\textwidth}{!}{%
\begin{tabular}{@{}cl|lll|lll|lll|lll|lll|lll@{}}
\toprule
\multicolumn{1}{l}{\textbf{}} &
  \textbf{} &
  \multicolumn{3}{c|}{2 months} &
  \multicolumn{3}{c|}{6 weeks} &
  \multicolumn{3}{c|}{1 month} &
  \multicolumn{3}{c|}{3 weeks} &
  \multicolumn{3}{c|}{2 weeks} &
  \multicolumn{3}{c}{1 week} \\
\multicolumn{1}{l}{Target} &
  Source &
  \multicolumn{1}{c}{FT} &
  \multicolumn{1}{c}{CG} &
  \multicolumn{1}{c|}{Ours} &
  \multicolumn{1}{c}{FT} &
  \multicolumn{1}{c}{CG} &
  \multicolumn{1}{c|}{Ours} &
  \multicolumn{1}{c}{FT} &
  \multicolumn{1}{c}{CG} &
  \multicolumn{1}{c|}{Ours} &
  \multicolumn{1}{c}{FT} &
  \multicolumn{1}{c}{CG} &
  \multicolumn{1}{c|}{Ours} &
  \multicolumn{1}{c}{FT} &
  \multicolumn{1}{c}{CG} &
  \multicolumn{1}{c|}{Ours} &
  \multicolumn{1}{c}{FT} &
  \multicolumn{1}{c}{CG} &
  \multicolumn{1}{c}{Ours} \\ \midrule
\multirow{4}{*}{WT01} &
  WT07 &
  \textbf{89.0} &
  87.9 &
  62.0 &
  \textbf{88.5} &
  86.7 &
  55.6 &
  77.5 &
  \textbf{86.1} &
  59.0 &
  \textbf{73.1} &
  57.3 &
  64.6 &
  74.0 &
  \textbf{79.4} &
  77.7 &
  74.7 &
  \textbf{78.8} &
  68.5 \\
 &
  WT02 &
  56.8 &
  64.4 &
  \textbf{76.3} &
  77.4 &
  \textbf{86.3} &
  75.5 &
  72.2 &
  \textbf{89.5} &
  75.6 &
  70.2 &
  \textbf{79.6} &
  77.0 &
  72.9 &
  \textbf{83.2} &
  81.8 &
  72.3 &
  76.7 &
  \textbf{78.0} \\
 &
  WT05 &
  83.1 &
  82.2 &
  \textbf{85.2} &
  80.9 &
  \textbf{88.3} &
  87.6 &
  73.7 &
  84.0 &
  \textbf{90.2} &
  75.1 &
  87.9 &
  \textbf{88.0} &
  79.0 &
  83.6 &
  \textbf{87.0} &
  75.2 &
  75.5 &
  \textbf{87.3} \\
 &
  WT04 &
  85.1 &
  \textbf{90.6} &
  83.2 &
  80.6 &
  \textbf{90.3} &
  88.5 &
  78.3 &
  84.0 &
  \textbf{88.9} &
  71.5 &
  \textbf{90.8} &
  88.5 &
  72.7 &
  88.9 &
  \textbf{89.8} &
  66.6 &
  78.4 &
  \textbf{89.5} \\ \midrule
\multirow{4}{*}{WT02} &
  WT01 &
  \textbf{80.6} &
  50.0 &
  64.1 &
  \textbf{83.7} &
  64.6 &
  64.2 &
  \textbf{84.3} &
  56.3 &
  61.3 &
  \textbf{73.9} &
  56.6 &
  62.2 &
  32.7 &
  42.5 &
  \textbf{50.4} &
  1.9 &
  41.8 &
  \textbf{44.7} \\
 &
  WT06 &
  62.2 &
  \textbf{71.7} &
  71.4 &
  \textbf{80.9} &
  69.0 &
  75.0 &
  \textbf{84.0} &
  80.0 &
  68.0 &
  74.9 &
  \textbf{77.1} &
  69.9 &
  40.7 &
  \textbf{64.7} &
  61.4 &
  9.1 &
  \textbf{62.1} &
  53.9 \\
 &
  WT05 &
  \textbf{82.1} &
  63.5 &
  63.2 &
  \textbf{80.8} &
  64.3 &
  64.2 &
  \textbf{74.5} &
  65.2 &
  61.4 &
  \textbf{77.7} &
  65.5 &
  55.3 &
  \textbf{82.6} &
  61.2 &
  55.7 &
  \textbf{71.3} &
  61.4 &
  50.1 \\
 &
  WT04 &
  \textbf{84.7} &
  64.8 &
  64.0 &
  \textbf{83.9} &
  67.0 &
  67.8 &
  \textbf{72.7} &
  67.1 &
  63.4 &
  \textbf{78.6} &
  70.7 &
  54.1 &
  \textbf{80.7} &
  63.4 &
  53.8 &
  \textbf{66.0} &
  55.1 &
  51.2 \\ \midrule
\multirow{6}{*}{WT03} &
  WT01 &
  36.1 &
  \textbf{86.5} &
  86.5 &
  76.6 &
  86.7 &
  \textbf{87.0} &
  56.9 &
  \textbf{84.8} &
  84.2 &
  76.8 &
  73.3 &
  \textbf{88.3} &
  67.1 &
  73.5 &
  \textbf{78.2} &
  17.1 &
  40.7 &
  \textbf{86.5} \\
 &
  WT06 &
  43.2 &
  \textbf{79.1} &
  59.1 &
  78.4 &
  \textbf{79.6} &
  51.3 &
  67.0 &
  \textbf{72.1} &
  66.4 &
  \textbf{75.2} &
  40.5 &
  66.9 &
  62.8 &
  \textbf{85.4} &
  57.6 &
  17.8 &
  41.7 &
  \textbf{59.9} \\
 &
  WT07 &
  53.2 &
  \textbf{86.8} &
  26.3 &
  71.7 &
  \textbf{85.6} &
  22.2 &
  \textbf{79.3} &
  48.7 &
  25.9 &
  \textbf{84.1} &
  46.0 &
  35.0 &
  76.3 &
  \textbf{89.3} &
  25.0 &
  54.8 &
  50.2 &
  \textbf{65.0} \\
 &
  WT02 &
  53.3 &
  \textbf{78.4} &
  47.2 &
  78.9 &
  \textbf{83.4} &
  46.2 &
  \textbf{86.5} &
  48.7 &
  52.7 &
  73.7 &
  \textbf{76.2} &
  61.2 &
  66.5 &
  \textbf{78.6} &
  46.3 &
  41.0 &
  52.5 &
  \textbf{73.6} \\
 &
  WT05 &
  \textbf{88.8} &
  72.4 &
  81.2 &
  \textbf{89.9} &
  85.5 &
  85.3 &
  76.9 &
  82.4 &
  \textbf{83.1} &
  78.9 &
  \textbf{85.6} &
  82.9 &
  \textbf{88.7} &
  73.2 &
  82.0 &
  \textbf{90.8} &
  61.6 &
  79.3 \\
 &
  WT04 &
  81.7 &
  77.8 &
  \textbf{91.2} &
  89.8 &
  79.3 &
  \textbf{91.6} &
  74.8 &
  65.9 &
  \textbf{87.7} &
  81.7 &
  84.6 &
  \textbf{85.6} &
  85.9 &
  70.0 &
  \textbf{86.9} &
  65.1 &
  48.7 &
  \textbf{77.4} \\ \midrule
\multirow{5}{*}{WT04} &
  WT01 &
  \textbf{79.8} &
  75.4 &
  79.5 &
  70.9 &
  \textbf{78.8} &
  72.7 &
  78.8 &
  \textbf{85.0} &
  71.8 &
  67.4 &
  \textbf{82.3} &
  64.6 &
  42.3 &
  54.5 &
  \textbf{55.3} &
  36.8 &
  43.4 &
  \textbf{59.8} \\
 &
  WT06 &
  \textbf{86.3} &
  82.3 &
  77.0 &
  \textbf{85.7} &
  72.3 &
  62.8 &
  \textbf{81.0} &
  80.5 &
  62.2 &
  64.3 &
  \textbf{86.7} &
  67.6 &
  45.3 &
  49.7 &
  \textbf{55.6} &
  42.3 &
  47.0 &
  \textbf{57.4} \\
 &
  WT07 &
  \textbf{88.6} &
  84.3 &
  86.8 &
  \textbf{92.3} &
  84.9 &
  74.4 &
  \textbf{89.9} &
  82.5 &
  72.6 &
  \textbf{83.3} &
  78.9 &
  71.9 &
  60.4 &
  66.3 &
  \textbf{69.9} &
  57.3 &
  61.3 &
  \textbf{64.6} \\
 &
  WT02 &
  88.5 &
  \textbf{92.3} &
  86.1 &
  81.1 &
  \textbf{89.6} &
  74.0 &
  87.8 &
  \textbf{91.5} &
  73.0 &
  82.8 &
  \textbf{85.4} &
  73.8 &
  63.3 &
  69.9 &
  \textbf{72.9} &
  52.4 &
  \textbf{61.5} &
  49.0 \\
 &
  WT05 &
  91.7 &
  \textbf{93.4} &
  95.3 &
  91.8 &
  \textbf{93.5} &
  90.2 &
  92.8 &
  \textbf{94.7} &
  88.4 &
  92.1 &
  \textbf{93.5} &
  86.5 &
  87.3 &
  61.1 &
  \textbf{87.6} &
  \textbf{93.6} &
  63.4 &
  76.0 \\ \midrule
\multirow{5}{*}{WT05} &
  WT01 &
  \textbf{65.1} &
  61.9 &
  53.7 &
  64.9 &
  51.6 &
  \textbf{72.9} &
  53.8 &
  \textbf{62.4} &
  52.5 &
  46.4 &
  45.2 &
  \textbf{49.7} &
  26.0 &
  33.2 &
  \textbf{50.0} &
  31.9 &
  19.5 &
  \textbf{67.4} \\
 &
  WT06 &
  65.8 &
  \textbf{78.3} &
  75.6 &
  63.5 &
  \textbf{76.6} &
  68.5 &
  54.8 &
  \textbf{70.7} &
  70.1 &
  50.0 &
  \textbf{80.8} &
  68.8 &
  34.1 &
  \textbf{78.7} &
  74.7 &
  13.1 &
  47.4 &
  \textbf{76.0} \\
 &
  WT07 &
  \textbf{87.4} &
  75.7 &
  64.3 &
  \textbf{83.5} &
  76.1 &
  65.8 &
  68.8 &
  \textbf{76.5} &
  63.5 &
  69.0 &
  \textbf{78.8} &
  68.0 &
  64.5 &
  47.9 &
  \textbf{67.8} &
  60.8 &
  52.1 &
  \textbf{67.5} \\
 &
  WT02 &
  \textbf{87.4} &
  81.8 &
  69.9 &
  \textbf{86.5} &
  80.7 &
  76.2 &
  83.2 &
  \textbf{86.1} &
  70.0 &
  76.5 &
  \textbf{88.9} &
  76.2 &
  71.8 &
  62.9 &
  \textbf{76.7} &
  49.4 &
  53.1 &
  \textbf{77.5} \\
 &
  WT04 &
  73.7 &
  \textbf{93.2} &
  75.3 &
  65.5 &
  \textbf{88.1} &
  85.7 &
  70.7 &
  73.4 &
  \textbf{79.7} &
  68.2 &
  83.5 &
  \textbf{84.0} &
  82.5 &
  79.5 &
  \textbf{83.0} &
  51.2 &
  57.6 &
  \textbf{81.8} \\ \midrule
\multirow{4}{*}{WT06} &
  WT07 &
  67.3 &
  66.2 &
  \textbf{78.6} &
  59.1 &
  60.8 &
  \textbf{79.2} &
  40.8 &
  \textbf{89.7} &
  78.6 &
  36.3 &
  49.6 &
  \textbf{78.6} &
  38.0 &
  51.0 &
  \textbf{78.3} &
  6.4 &
  10.9 &
  \textbf{63.6} \\
 &
  WT02 &
  63.7 &
  76.7 &
  \textbf{78.0} &
  60.5 &
  73.6 &
  \textbf{78.6} &
  42.5 &
  74.5 &
  \textbf{76.6} &
  39.3 &
  47.0 &
  \textbf{74.9} &
  32.5 &
  40.1 &
  \textbf{77.2} &
  6.1 &
  6.9 &
  \textbf{54.9} \\
 &
  WT05 &
  74.9 &
  \textbf{94.9} &
  78.3 &
  42.9 &
  57.8 &
  \textbf{78.6} &
  20.7 &
  54.0 &
  \textbf{77.7} &
  18.3 &
  75.7 &
  \textbf{77.7} &
  19.5 &
  70.0 &
  \textbf{77.7} &
  6.2 &
  5.7 &
  \textbf{57.4} \\
 &
  WT04 &
  79.4 &
  \textbf{94.9} &
  73.6 &
  59.7 &
  \textbf{76.5} &
  74.6 &
  28.1 &
  72.3 &
  \textbf{74.9} &
  15.9 &
  \textbf{75.8} &
  74.1 &
  21.5 &
  \textbf{78.9} &
  72.3 &
  6.0 &
  7.1 &
  \textbf{47.7} \\ \bottomrule
\end{tabular}%
}
\end{table}
\end{document}